\documentclass[10pt,twocolumn,letterpaper]{article}

\usepackage[accsupp]{axessibility}
\usepackage[pagenumbers]{cvpr} 

\usepackage[dvipsnames]{xcolor}

\newcommand{\minisection}[1]{\vspace{2mm}\noindent{\textbf{#1}}}

\usepackage{xcolor}
\usepackage{url}
\usepackage{pifont}
\usepackage{textcomp, gensymb}
\usepackage{multirow}

\newcommand{\red}[1]{{\textcolor{red}{#1}}}
\newcommand{\blue}[1]{{\textcolor{blue}{#1}}}

\definecolor{applegreen}{rgb}{0.55, 0.71, 0.0}
\newcommand{\applegreen}[1]{{\textcolor{applegreen}{#1}}}
\definecolor{cvprblue}{rgb}{0.21,0.49,0.74}
\newcommand{\cmark}{\ding{51}}%
\usepackage[pagebackref,breaklinks,colorlinks,citecolor=cvprblue]{hyperref}

\def\paperID{15511} 
\def\confName{XXXXX}
\def\confYear{XXXXXX}

\title{EFHQ: Multi-purpose \underline{E}xtremePose-\underline{F}ace-\underline{HQ} dataset}

\author{
Trung Tuan Dao$^{1,*}$ \quad Duc Hong Vu$^{1,2,*}$ \quad Cuong Pham$^{1,3}$ \quad Anh Tran$^{1}$ \\ 
\small{\textsuperscript{1}VinAI Research, Vietnam \quad \textsuperscript{2}Miami University \quad 
\textsuperscript{3}Posts \& Telecommunications Inst. of Tech., Vietnam
}\\
\texttt{\scriptsize \{v.trungdt21, v.ducvh5, v.cuongpv11, v.anhtt152\}@vinai.io} \quad 
\texttt{\scriptsize vudh3@miamioh.edu} \quad
\texttt{\scriptsize cuongpv@ptit.edu.vn   }\\
\small{\textsuperscript{*}Equal contribution}
}

\begin{document}
\twocolumn[{%
\renewcommand\twocolumn[1][]{#1}%
\maketitle
\begin{center}
  \includegraphics[width=.85\textwidth]{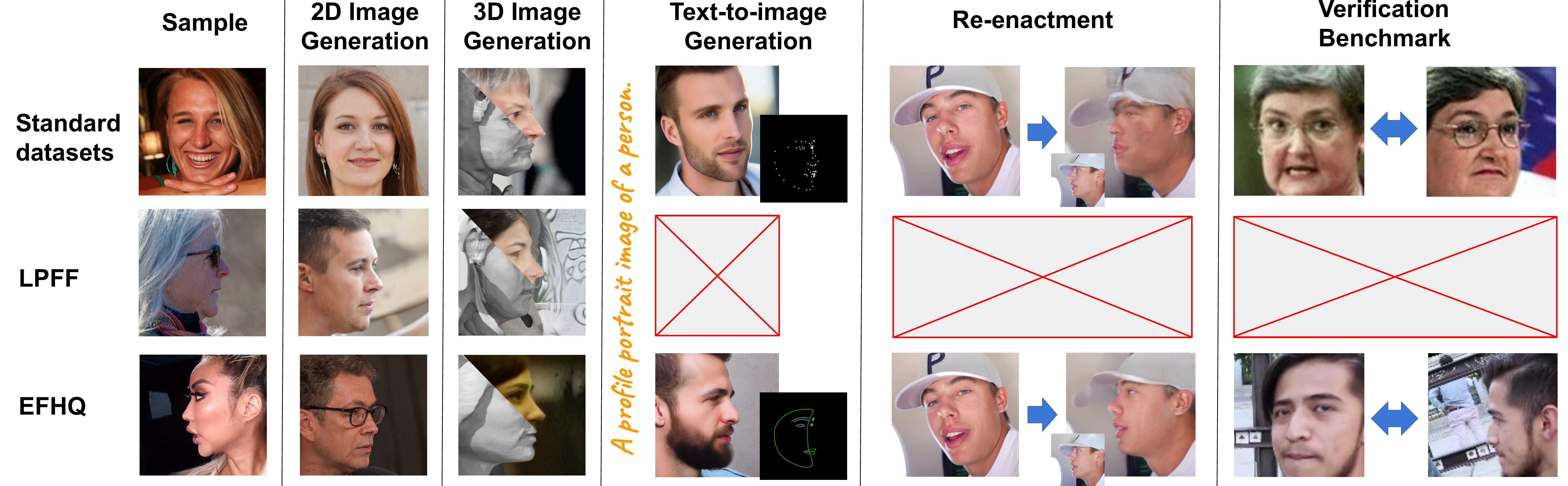}
  {\captionof{figure}{\textbf{Benefits of our proposed dataset (EFHQ).} Standard large-scale facial datasets have most images at near frontal views, causing inferior performance of trained models on downstream tasks when dealing with extreme head poses. For instance, the trained 2D image generators and text-to-image ones often produce only near frontal faces, while the 3D face generators and face reenactment methods often show distorted outputs at profile views. The recently proposed dataset LPFF \cite{lpff} partially handles that issue by providing complementary images at extreme head poses for only 2D and 3D image generation tasks. Our proposed dataset EFHQ provides high-quality extreme-pose images to complement a wide range of face-related tasks. It supports 2D and 3D image generation, with generally better diversity than LPFF. EFHQ also helps correct the outputs of text-to-image generation and face reenactment at extreme views. Finally, EFHQ provides a more challenging pose-based face verification benchmark to better assess the quality of face recognition networks.}
  \label{fig:teaser}}
\end{center}%
}]
\maketitle
\begin{abstract}
The existing facial datasets, while having plentiful images at near frontal views, lack images with extreme head poses, leading to the downgraded performance of deep learning models when dealing with profile or pitched faces. This work aims to address this gap by introducing a novel dataset named Extreme Pose Face High-Quality Dataset (EFHQ), which includes a maximum of 450k high-quality images of faces at extreme poses. To produce such a massive dataset, we utilize a novel and meticulous dataset processing pipeline to curate two publicly available datasets, VFHQ and CelebV-HQ, which contain many high-resolution face videos captured in various settings. Our dataset can complement existing datasets on various facial-related tasks, such as facial synthesis with 2D/3D-aware GAN, diffusion-based text-to-image face generation, and face reenactment. Specifically, training with EFHQ helps models generalize well across diverse poses, significantly improving performance in scenarios involving extreme views, confirmed by extensive experiments. Additionally, we utilize EFHQ to define a challenging cross-view face verification benchmark, in which the performance of SOTA face recognition models drops 5-37\% compared to frontal-to-frontal scenarios, aiming to stimulate studies on face recognition under severe pose conditions in the wild.
\end{abstract}    
\section{Introduction}
\label{sec:intro}
The human face is a central focus in computer vision, with extensive research dedicated to tasks like detection, recognition, generation, and manipulation \cite{eigenface1,eigenface2}. Numerous large-scale facial datasets\cite{lfw,ytf,agedb,ms1m,megaface2,webface260m,celeba,stylegan,celebvhq,vfhq} have facilitated the face-related studies. Along with deep learning techniques, many tasks have made a considerable leap in performance in recent years. For instance, face recognition systems can recognize near-frontal faces in the wild with near-perfect accuracy \cite{huang2020curricularface,kim2022adaface}. Facial generative models, such as \cite{stylegan2ada,eg3d,controlnet}, can synthesize realistic images indistinguishable from real faces. Face reenactment systems, including \cite{tps,lia,dam}, can animate an input face based on poses and expressions from any driver video.

Despite impressive results, the mentioned models produce inferior results when dealing with faces at extreme head poses. As 
illustrated in \cref{fig:teaser}, faces synthesized by 3D face generation or face reenactment models are often distorted at profile views, due to
the scarcity of extreme pose images in training datasets. Current extensive datasets for face generation leverage portrait images from Google or stock photo repositories like Flickr, which feature images with upfront views and minimal facial rotations. Face reenactment methods leverage video datasets such as Vox-Celeb \cite{vox1}, primarily featuring frames with frontal faces. The instability of pre-processing tools, particularly face alignment modules when handling extreme pose images, 
further reduces the proportion of those images in the existing datasets. This issue is critical since faces with large head rotations are common in real life, and dealing with them is unavoidable in practical systems.

This paper aims to fill that void by proposing a novel large-scale dataset of extreme-pose and high-quality facial images called ExtremePose-Face-HQ, or EFHQ for short. Our ambition is to build an ``one-for-all'' dataset that can supplement existing datasets on all mentioned face-related tasks. To achieve that goal, the dataset needs all the following properties: (1) extreme-pose, (2) large-scale, (3) high-quality, (4) in-the-wild, (5) having multiple images for each subject and having subject identity annotated. Given those requirements, we utilize a novel and meticulous dataset processing pipeline to filter and sample frames from two publicly available datasets, including VFHQ \cite{vfhq} and CelebV-HQ \cite{celebvhq}, which contain high-resolution face videos captured in various settings. This results in a dataset with 450k high-quality images of extreme poses, which are combined with frontal images of the same subject to use as image pairs. These image pairs can support face reenactment and face recognition tasks, while the extreme-pose images alone can complement the existing datasets of face generation training. Note that there is a recent attempt \cite{lpff} to build a large-pose facial dataset similar to ours. However, that work relies on images crawled from Flickr, thus having limited size (19k images) and missing identity information. Hence, its applications are restricted, covering only several face-generation tasks, unlike our large-scale and multi-purpose dataset.

We perform a comprehensive set of experiments to validate the effectiveness of EFHQ across three distinct face generation sub-tasks and face reenactment scenarios. Our dataset, EFHQ, when combined with standard training sets, significantly enhances the quality of synthesized images in highly profile or pitched views while maintaining synthesis quality in frontal views. EFHQ also provides a more challenging pose-centric face verification benchmark. It reveals the vulnerability of many state-of-the-art face recognition networks when dealing with extreme-pose images, with TAR@FAR=1e-3 scores dropping 5-37\%.

In summary, our contributions include (1) a large-scale, high-quality, in-the-wild, extreme-pose dataset (EFHQ) designed to complement various face-related tasks, (2) a novel and meticulous dataset processing pipeline ensuring the creation and quality of the EFHQ dataset, and (3) a series of experiments, with plan-to-release models, demonstrating the usefulness of EFHQ in enhancing performance on a wide range of tasks when dealing with faces at extreme rotations. The final dataset will be publicly available as metadata files to not violate the two original dataset's licenses, alongside the pretrained networks.


\begin{table*}[ht]
\centering
\resizebox{0.8\linewidth}{!}{
\begin{tabular}{l|c|ccc|ccc}
\hline
\multicolumn{1}{c|}{\textbf{}} &
  \textbf{} &
  \multicolumn{3}{c|}{\textbf{Attributes}} &
  \multicolumn{3}{c}{\textbf{Ready-to-use tasks}} \\ \cline{2-8} 
\multicolumn{1}{c|}{\textbf{Dataset}} &
  \textbf{\#images} &
  \multicolumn{1}{c|}{\textbf{\begin{tabular}[c]{@{}c@{}}Multiple images \\ per ID\end{tabular}}} &
  \multicolumn{1}{c|}{\textbf{\begin{tabular}[c]{@{}c@{}}High \\ Quality\end{tabular}}} &
  \textbf{\begin{tabular}[c]{@{}c@{}}Profile \\ View\end{tabular}} &
  \multicolumn{1}{c|}{\textbf{\begin{tabular}[c]{@{}c@{}}Face \\ Verification\end{tabular}}} &
  \multicolumn{1}{c|}{\textbf{\begin{tabular}[c]{@{}c@{}}Face \\ Synthesis\end{tabular}}} &
  \textbf{\begin{tabular}[c]{@{}c@{}}Face \\ Reenactment\end{tabular}} \\ \hline
\textbf{AgeDB} &
  12K &
  \multicolumn{1}{c|}{ \cmark} &
  \multicolumn{1}{c|}{} &
   &
  \multicolumn{1}{c|}{ \cmark} &
  \multicolumn{1}{c|}{} &
   \\
\textbf{CPLFW} &
  12K &
  \multicolumn{1}{c|}{ \cmark} &
  \multicolumn{1}{c|}{} &
   \cmark &
  \multicolumn{1}{c|}{ \cmark} &
  \multicolumn{1}{c|}{} &
   \\
\textbf{CFP-FP} &
  14K &
  \multicolumn{1}{c|}{ \cmark} &
  \multicolumn{1}{c|}{} &
   \cmark   &
  \multicolumn{1}{c|}{ \cmark} &
  \multicolumn{1}{c|}{} &
   \\
\textbf{IJB-x} &
  138K &
  \multicolumn{1}{c|}{ \cmark} &
  \multicolumn{1}{c|}{} &
   \cmark &
  \multicolumn{1}{c|}{ \cmark} &
  \multicolumn{1}{c|}{} &
   \\ \hline
\textbf{CelebA-HQ} &
  30K &
  \multicolumn{1}{c|}{} &
  \multicolumn{1}{c|}{ \cmark} &
   &
  \multicolumn{1}{c|}{} &
  \multicolumn{1}{c|}{ \cmark} &
   \\
\textbf{FFHQ} &
  70K &
  \multicolumn{1}{c|}{} &
  \multicolumn{1}{c|}{ \cmark} &
   &
  \multicolumn{1}{c|}{} &
  \multicolumn{1}{c|}{ \cmark} &
   \\
\textbf{LPFF} &
  19K &
  \multicolumn{1}{c|}{} &
  \multicolumn{1}{c|}{ \cmark} &
   \cmark &
  \multicolumn{1}{c|}{} &
  \multicolumn{1}{c|}{ \cmark} &
   \\ \hline
\textbf{CelebV-HQ} &
  6M &
  \multicolumn{1}{c|}{ \cmark} &
  \multicolumn{1}{c|}{ \cmark} &
   \cmark &
  \multicolumn{1}{c|}{ } &
  \multicolumn{1}{c|}{ \cmark} &
    \\
\textbf{VFHQ} &
  3M &
  \multicolumn{1}{c|}{ \cmark} &
  \multicolumn{1}{c|}{ \cmark} &
   \cmark &
  \multicolumn{1}{c|}{ } &
  \multicolumn{1}{c|}{ \cmark} &
   \cmark \\
\textbf{VoxCeleb1} &
  4M &
  \multicolumn{1}{c|}{ \cmark} &
  \multicolumn{1}{c|}{} &
  {} &
  \multicolumn{1}{c|}{} &
  \multicolumn{1}{c|}{\cmark} &
   \cmark \\ \hline
\textbf{Ours} &
  450,538 &
  \multicolumn{1}{c|}{ \cmark} &
  \multicolumn{1}{c|}{ \cmark} &
   \cmark &
  \multicolumn{1}{c|}{ \cmark} &
  \multicolumn{1}{c|}{ \cmark} &
   \cmark \\
\textbf{FFHQ+Ours*} &
  112,671 &
  \multicolumn{1}{c|}{} &
  \multicolumn{1}{c|}{ \cmark} &
   \cmark &
  \multicolumn{1}{c|}{} &
  \multicolumn{1}{c|}{ \cmark} &
   \\
\textbf{Vox+Ours*} &
  4.5M &
  \multicolumn{1}{c|}{ \cmark} &
  \multicolumn{1}{c|}{} &
   \cmark &
  \multicolumn{1}{c|}{} &
  \multicolumn{1}{c|}{} &
   \cmark \\ \hline
\end{tabular}
}
\vspace{-2mm}
\caption{\textbf{Representative comparison} of key attributes and supported tasks across existing face datasets and our proposed EFHQ dataset. EFHQ provides large-scale, high-quality images with multiple views per identity, enabling diverse face-related tasks, including synthesis, reenactment, and verification. Combining EFHQ with existing datasets like FFHQ or VoxCeleb can augment available data for selected tasks. Asterisk (*) indicates a combined dataset with a subsampled version of EFHQ.}
\label{tab:introtable}
\vspace{-4mm}

\end{table*}

\section{Related Work}
\label{sec:related_work}

\minisection{Synthetic Face Generation.} 
Existing 2D-based models like StyleGAN \cite{stylegan, stylegan2ada, stylegan3} and 3D-aware models like EG3D \cite{eg3d} demonstrate impressive fidelity, especially for near-frontal views. Nevertheless, their performance significantly degrades when faced with extreme views, primarily due to insufficient training data for such scenarios. Datasets like FFHQ \cite{stylegan} exhibit a notable bias toward near-frontal poses, heavily limiting the range of poses synthesized by current 2D and 3D models despite the 3D-aware generators' ability to handle arbitrary poses.

Diffusion-based text-to-image models \cite{stablediffusion,dalle2} can synthesize high-quality images from simple text inputs. However, these models lack mechanisms to control the generated pose. Consequently, the generated facial images remain in a frontal view despite terms like ``profile" or ``looking-up" in the input prompts.
ControlNet\cite{controlnet} proposes a finetuning approach allowing these models to take additional context as conditional input. Namely, one can use landmarks to guide the generated face to the desired pose. However, the scarcity of profile-view data in training affects the model's performance, particularly in extreme angles. Consequently, effectively managing the long-tailed distribution of these extreme views becomes necessary to address this issue.

\minisection{Face Reenactment.} 
Face reenactment involves transferring facial expressions and movements from a driving image to a source image. Recent advancements claim high-quality results, especially for near-frontal poses \cite{dam, tps, lia, dpe}. However, pose collapse persists, particularly in keypoint-based models \cite{dam, tps}. Common datasets for this task are VoxCeleb1 \cite{vox1} and HDTF \cite{hdtf}, which primarily feature frontal talking face videos. Despite recent potentially high-quality datasets for reenactment such as VFHQ \cite{vfhq} and CelebV-HQ \cite{celebvhq}, the frontal-view bias persists. As a result, facial reenactment methods often produce artifacts and lower-quality outputs when faced with significant rotations.

\minisection{Face Verification.} Despite near-perfect accuracy on benchmark datasets\cite{agedb,ijba,ijbb,ijbc,lfw}, state-of-the-art face verification models\cite{deng2019arcface,huang2020curricularface,kim2022adaface} fall short in unconstrained environments due to a lack of pose diversity. While CPLFW\cite{cplfw} introduces 6000 cross-pose pairs, its small size and mix of pose samples hinder analyzing performance on specific pose scenarios such as frontal-frontal, frontal-profile, and profile-profile. More granular benchmarking on diverse pose pairs is crucial to identify failure modes and drive progress in robust face verification across poses.


\minisection{Extreme Pose Face Dataset.} \cref{tab:introtable} summarizes existing datasets related to face verification, synthesis, and reenactment, showing that existing face datasets have limitations for multi-task use. Recent efforts like LPFF\cite{lpff} made progress by crawling from Flickr and processing for diverse views, resulting in a dataset of 19,590 images. The dataset could complement high-quality datasets like FFHQ\cite{stylegan} for generative models. However, due to its limited scale comprising only 20\% of the combined dataset, the unbalanced distribution between frontal and profile view still needs further improvements.
Additionally, since the images are from image hosting services, the dataset fails to provide the identity information necessary for other facial applications.

\begin{figure*}[ht!]
\centering
    \vspace{-2mm}
\includegraphics[width=.8\linewidth]{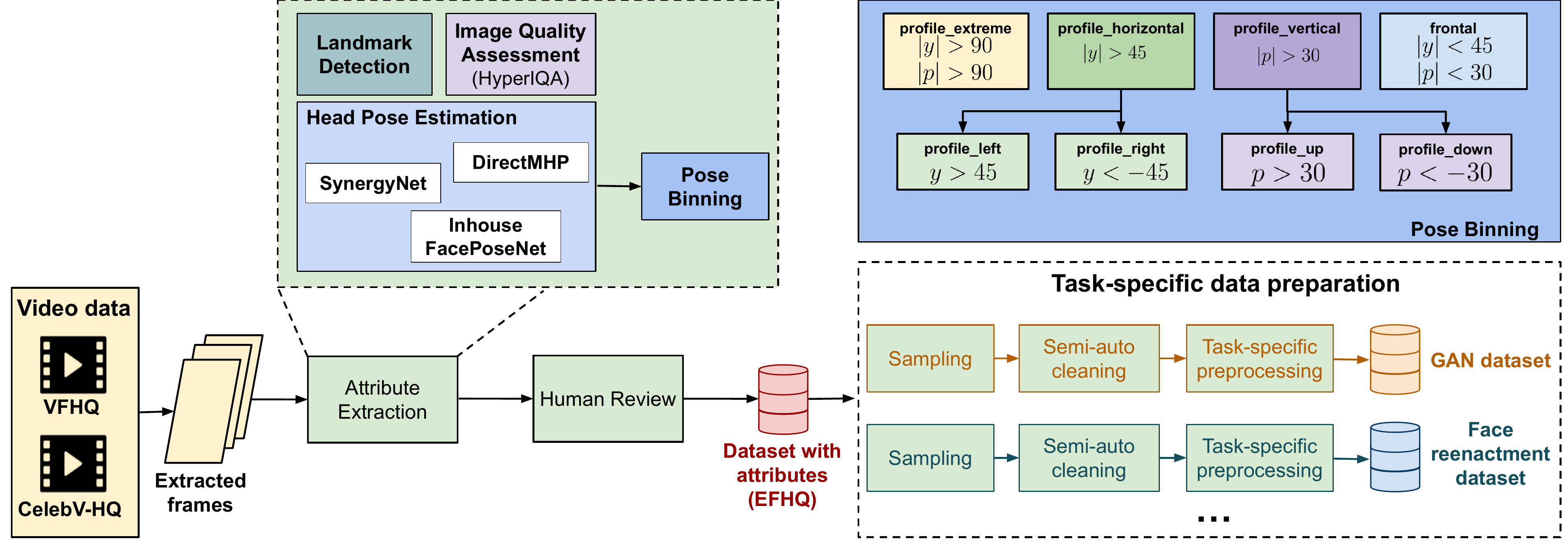}
    \vspace{-2mm}
\caption{\textbf{The pipeline of EFHQ dataset creation.} Starting with high-quality videos from the VFHQ\cite{vfhq} and CelebV-HQ\cite{celebvhq} datasets, single-frame attributes are extracted then manually reviews. Task-specific preprocessing is then applied to generate specialized versions of the dataset for tasks such as face generation, reenactment, and verification.}
    \vspace{-2mm}
\label{fig:pipeline}
\end{figure*}
\section{Dataset Process}
\label{sec:dataset_intro}

In this section, we discuss the process of producing our novel dataset EFHQ. To fulfill our goals, we carefully curated frames from the two recent facial video datasets, VFHQ\cite{vfhq} and CelebV-HQ\cite{celebvhq}
, as seen in \cref{fig:pipeline}.

\subsection{Attributes Preparation}

\minisection{Attributes Extraction.} To curate accurate facial poses and high-quality frames from existing datasets, we first extract (1) face bounding boxes, (2) facial landmarks (5/68 keypoints), (3) image quality score, and (4) face identity. For bounding boxes and landmarks, we employ the popular RetinaFace\cite{retinaface} and SynergyNet\cite{synergynet}. We use HyperIQA \cite{hyperiqa} to grade image quality. For VFHQ, we only generate 68 landmarks and image quality scores since bounding boxes and 5 keypoints are provided. After extracting these attributes, we match them with existing annotations using an IoU-based Hungarian matching algorithm \cite{kuhn1955hungarian}. Regarding identity, VFHQ-wise, we source labels from the annotations. As CelebV-HQ lacks individual identity labels within the video, we use\cite{lightface} to identify the identities associated with each bounding box.

\begin{figure}[ht!]
\centering
\includegraphics[width=0.9\linewidth]{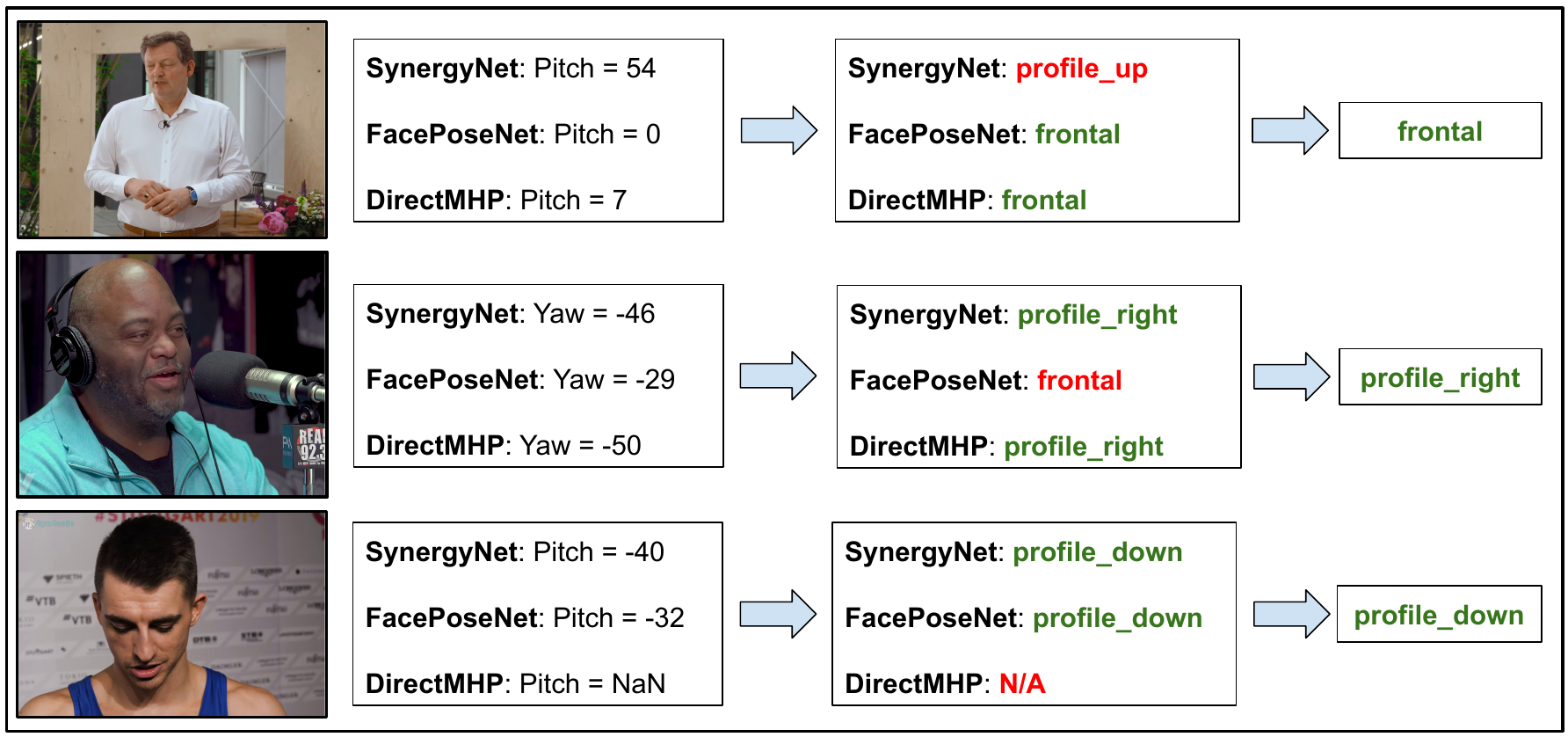}
    \vspace{-2mm}
\caption{\textbf{Example cases} where a pose estimator fails to categorize the sample to the correct bin.}
\label{fig:ensemble_example}
    \vspace{-4mm}
\end{figure}

Due to the instability of head pose estimators when handling extreme-pose images, we strategically apply multiple pose estimators relying on various methodologies: (1) SynergyNet\cite{synergynet}, a 3DMM-based model, (2) DirectMHP\cite{mhp}, a landmark-free joint head detector and pose estimator, and (3) an in-house FacePoseNet\cite{faceposenet} improved through extensive data augmentation focused on extreme poses and extra training data. We then categorize each estimated pose into a hierarchical binning scheme illustrated in \cref{fig:pipeline}. Next, we gather bin predictions and perform majority voting to arrive at consensus labels. When no agreement emerges, we put it into the $5^{th}$ bin for ``confusing'' cases. As depicted in \cref{fig:ensemble_example}, this ensemble approach provides robustness in cases where a single estimator incorrectly categorizes a sample. Please refer to the supplementary for statistics of the extracted bins and the hyperparameters for each model used for attribute extraction.

\minisection{Annotation Review.} Next, we manually review the results to ensure high-quality labels. Given the large dataset, we developed a graphical user interface tool to streamline the review process. Our focus is verifying the binning annotations, which most directly impact label accuracy. Concurrently, we randomly subsample the data based on pose angle and image quality to validate facial landmark quality further and discard cases of landmark prediction failure.

\subsection{EFHQ dataset}
Our final EFHQ dataset comprises up to 450k frames with extreme poses extracted from approximately 5,000 clips.
Most of the clips in our dataset include at least one frame with a frontal face and multiple frames with extreme pose angles. We also include extreme pose-only clips, representing profile-to-profile pose transfer cases.



\begin{figure}[ht!]
\centering
\includegraphics[width=.95\linewidth]{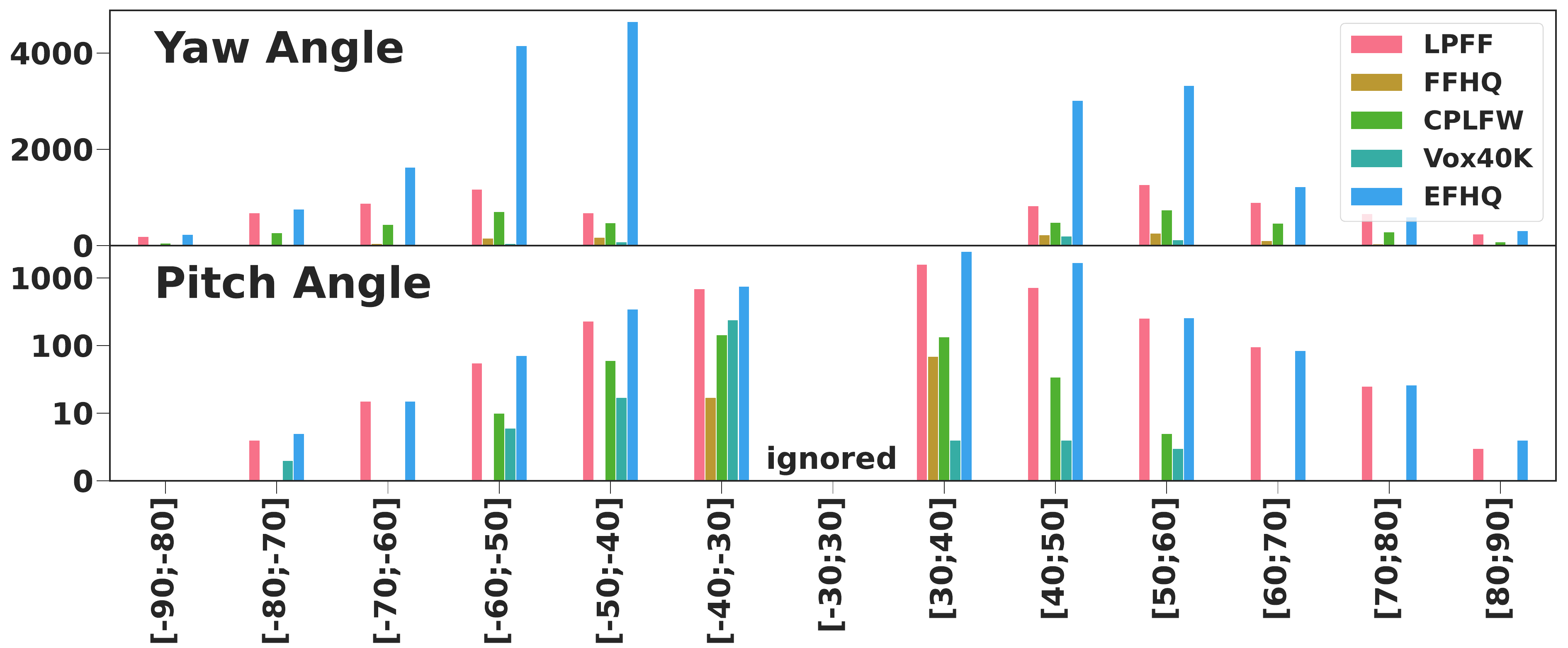}
\vspace{-2mm}
\caption{\textbf{Pose distribution} comparison between our sampled dataset and other datasets, including FFHQ\cite{stylegan}, LPFF\cite{lpff}, CPLFW\cite{cplfw} and 40K random samples from VoxCeleb1 \cite{vox1}. Our sampled dataset demonstrates greater pose diversity, with increased sample counts across high angle bins.}
\label{fig:dist_compare}
\vspace{-5mm}
\end{figure}

\subsection{Dataset for subtasks}

\minisection{Supplementary datasets for face generation.}
    To address FFHQ's pose distribution gap, we compile a dataset encompassing diverse poses, varied identities, and FFHQ-comparable image quality from EFHQ.
    Additionally, we also embed an image brightness filter\cite{brightness} in the sampling process, given that HyperIQA doesn't explicitly assess this aspect. 
    The sampled images are then processed using the same pipeline in \cite{stylegan2, eg3d} and manually reviewed. The aim is to improve coverage of under-represented poses without degrading frontal performance of trained generation models. This yields a dataset of 42,671 images with equitably distributed poses as visualized in \cref{fig:dist_compare}. Compared to the original FFHQ distribution, yaw and pitch distributions are significantly improved.

    Additionally, for 3D-aware GAN subtasks, we extract camera parameters to serve as conditional pose information, following \cite{eg3d}. We convert these parameters into yaw, pitch angles to validate against our existing annotations. When disagreements emerge, we re-examine the images and filter if needed, to ensure accurate, quality labels.

    For diffusion-based text-to-image generation, a specialized dataset is curated to refine Stable Diffusion models\cite{stablediffusion} by accommodating landmark conditional input through ControlNet\cite{controlnet}. Based on our 3D-aware face generation dataset, we integrate landmark-based conditional images and tailored text prompts for each image. To capture fine facial nuances, we extract 478 facial landmarks and draw the condition image with connected edges between matched keypoints, following Mediapipe framework\cite{mediapipe}, better-conveying face expression and pose. For text prompts, detailed facial attributes like gender, race, and emotion are derived from existing labels or inferred via BLIP-2 pretrained captioning model \cite{blip2} and a face attribute estimator \cite{serengil2021lightface}. The resulting prompt adheres to a structured format: ``A profile portrait image of a [emotion] [race] [gender]."

\minisection{Supplementary dataset for face reenactment.} 
As EFHQ is crafted, we can employ the entire dataset to complement established training datasets, e.g., VoxCeleb1 \cite{vox1}. We also built an extra evaluation set with 1200 EFHQ clips.

\minisection{Benchmarking dataset for face verification.}
    We curate a dataset covering three distinct scenarios: frontal-to-frontal, frontal-to-profile, and profile-to-profile. To enable rigorous benchmarking, we sample 10,000 pairs each for both negative and positive cases per scenario, resulting in a balanced benchmark dataset containing 60,000 image pairs. Images follow the same established preprocessing pipelines in \cite{deng2019arcface}. Next, we filter misaligned images with RetinaFace\cite{retinaface}, randomly review, and replace equivalent samples from our diverse corpus, if needed. Finally, to simulate varying image quality, we randomly applied downscaling and compression to samples in the dataset.


\section{Face Generation Subtask}
\label{sec:facegen}
\begin{table}
  \centering
  \setlength{\tabcolsep}{8pt}
    \begin{tabular}{lcll}
    \toprule
    Model & Reference Dataset & FID$\downarrow$ & Recall$\uparrow$ \\
    \midrule
    $G^{\text{FFHQ}}$ & FFHQ & 2.84 & 0.49 \\
    \midrule
    $G^{\text{FFHQ+LPFF}}$ & FFHQ+LPFF & \textbf{3.43} & 0.44 \\
    $G^{\text{FFHQ+EFHQ}}$ & FFHQ+EFHQ & 3.44 & \textbf{0.46} \\    
    \midrule
    $G^{\text{FFHQ+LPFF}}$ & FFHQ+EFHQ & 8.26 & 0.43 \\
    $G^{\text{FFHQ+EFHQ}}$ & FFHQ+LPFF & \textbf{7.04} & \textbf{0.44} \\    
    \midrule
    $G_c^{\text{FFHQ+EFHQ}}$ & FFHQ+EFHQ & 3.33 & 0.46 \\
    $G_c^{\text{FFHQ+EFHQ}}$ & FFHQ & 3.12 & 0.44 \\
    $G_c^{\text{FFHQ+EFHQ}}$ & FFHQ+LPFF & 4.47 & 0.42 \\
    \bottomrule
  \end{tabular}
    \vspace{-2mm}
  \caption{\textbf{StyleGAN2-ADA models trained on different datasets.} Lower FIDs indicate better fidelity, while higher Recalls indicate better diversity. The second and third blocks show comparisons between unconditional models trained on FFHQ+LPFF and FFHQ+EFHQ when evaluating on the same and cross-dataset settings, with the better metrics in \textbf{bold}.}
  \label{tab:stylegan}
\vspace{-6mm}
\end{table}

\begin{figure}[ht]
\centering
\begin{subfigure}{0.402\textwidth}
    \includegraphics[width=\linewidth]{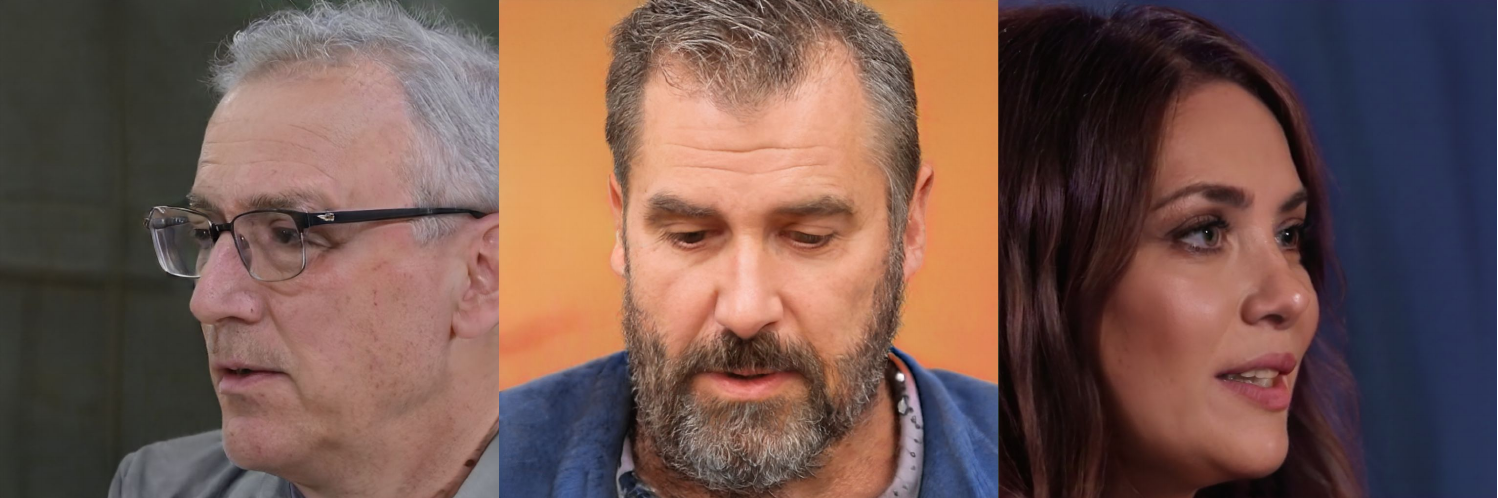}
    \label{fig:stylegan_pose}
    \vspace{-4mm}
    \caption{Samples from the model trained with FFHQ+EFHQ.}
\end{subfigure}
~
\centering
\begin{subfigure}{0.4\textwidth}
    \includegraphics[width=\linewidth]{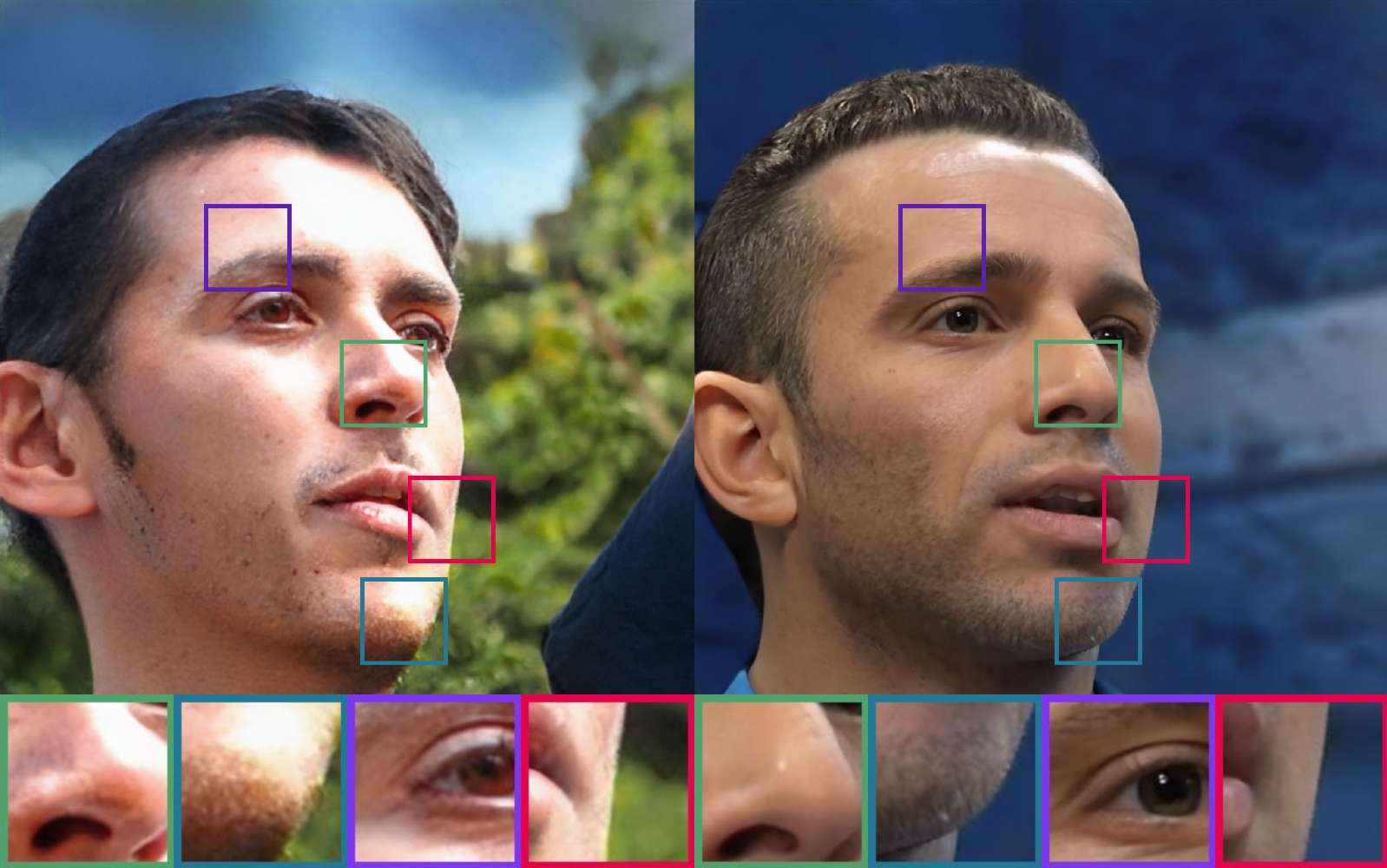}
    \vspace{-4mm}
    \caption{Profile-view comparison of LPFF (left) and EFHQ (right).}
    \label{fig:stylegan_compared}
\end{subfigure}
    \vspace{-2mm}
\caption{\textbf{Qualitative result and comparison} of generated samples, with truncation $\psi=0.7$, from StyleGAN2-ADA training with other dataset and ours.}
\label{fig:stylegan_big}
    \vspace{-6mm}
\end{figure}

\subsection{StyleGAN}
\minisection{Training details.}
    We train StyleGAN2-ADA\cite{stylegan2ada} models from scratch after combining FFHQ with our own curated dataset (denoted FFHQ+EFHQ) at $1024\times1024$ resolution.
We also train a conditional version, allowing direct control over frontal versus profile synthesis and enhancing model performance. Specifically, we categorize the dataset into two classes: ``frontal" and ``profile" based on pose angles. All models are trained for 6 days on 8 Nvidia A100 GPUs.

\minisection{Evaluation protocol.} 
    We utilize the Frechet Inception Distance\cite{fid} (FID) metric and improved Recall\cite{recall} to quantify the generation fidelity and diversity. Regarding the two unconditional models trained with the supplemented FFHQ+LPFF and FFHQ+EFHQ, we evaluate each model with both of these datasets. For conditional models, we evaluate performance on FFHQ, FFHQ+LPFF, and FFHQ+EFHQ to demonstrate the broader coverage of the model trained with our proposed dataset. For FFHQ, we generate 50K images with the ``frontal'' label, whereas for the other two datasets, we generate 50K following their label distributions. This comprehensive quantitative analysis examines the fidelity and variety of the generated images.

\minisection{Experimental results.} 	
    We compared the performance of our proposed unconditional models ($G^{\text{FFHQ+EFHQ}}$), against models trained on the FFHQ and FFHQ+LPFF datasets ($G^{\text{FFHQ}}$ and $G^{\text{FFHQ+LPFF}}$). Models trained conditionally are specifically subscripted, e.g., $G_{c}^{\text{FFHQ+EFHQ}}$. The qualitative results in \cref{fig:stylegan_big} illustrate that our models generate high-quality frontal faces comparable to the FFHQ model while producing realistic and varied profile faces. Additionally, our method has less noise in extreme poses than the FFHQ+LPFF model. Specifically, \cref{fig:stylegan_compared} shows LPFF's generated face contains more noise and less photorealistic details than our model, such as the pixelated eye and nose patches. Meanwhile, ours achieves greater realism with smoother skin, well-defined features, noise reduction, and natural lighting. The quantitative results in \cref{tab:stylegan} show that our unconditional model achieves competitive FID scores and improved Recall to other models, indicating similar fidelity but greater sample pose diversity. In the cross-examination (third block of \cref{tab:stylegan}), $G^{\text{FFHQ+EFHQ}}$ has both better FID and improved Recall on the opposite dataset than $G^{\text{FFHQ+LPFF}}$. Finally, our conditional model demonstrates substantially improved FID scores over multiple reference datasets, highlighting its versatility.

\begin{table}
  \centering
\resizebox{\columnwidth}{!}{
  \begin{tabular}{@{}lcccll@{}}
    \toprule
    Model                   & Reference Dataset  & FID$\downarrow$ & Pose$\downarrow$ & Depth$\downarrow$ & ID$\uparrow$   \\
    \midrule
    $E_{bug}^{\text{FFHQ}}$ & FFHQ/Trainset & 4.8             & $5e^{-4}$        & 0.234             & \underline{0.74} \\
    $E^{\text{FFHQ}}$       & FFHQ/Trainset & \underline{\red{3.7}}             & $5e^{-4}$        & 0.288             & \underline{0.74} \\
    \midrule
    $E^{\text{FFHQ+LPFF}}$  & FFHQ     & 6.2            & $6e^{-4}$        & 0.316             & \red{\textbf{0.77}} \\
    $E^{\text{FFHQ+EFHQ}}$  & FFHQ     & \textbf{3.9}             &  \red{$\boldsymbol{4e^{-4}}$}        & \red{\textbf{0.224}}             & \red{\textbf{0.77}} \\
    \midrule
    $E^{\text{FFHQ+LPFF}}$  & Trainset & 7.0             & $6e^{-4}$        & 0.316             & \textbf{0.65} \\
    $E^{\text{FFHQ+EFHQ}}$  & Trainset & \textbf{4.6}             & \underline{$\boldsymbol{4e^{-4}}$}        & \underline{\textbf{0.223}}             & \textbf{0.65} \\
    \bottomrule
  \end{tabular}
  }
    \vspace{-2mm}
  \caption{\textbf{EG3D models trained on different datasets.} In the second and third blocks, we compare models trained on FFHQ+LPFF and FFHQ+EFHQ with the testset is either FFHQ or their training datasets; the better metrics are in \textbf{bold}. The best metrics on the FFHQ testset are in \red{red}, while the best metrics when evaluating on training dataset are \underline{underlined}.}
  \label{tab:eg3d}
\vspace{-4mm}
\end{table}


\begin{figure}[ht!]
\centering
\includegraphics[keepaspectratio,width=0.95\linewidth]{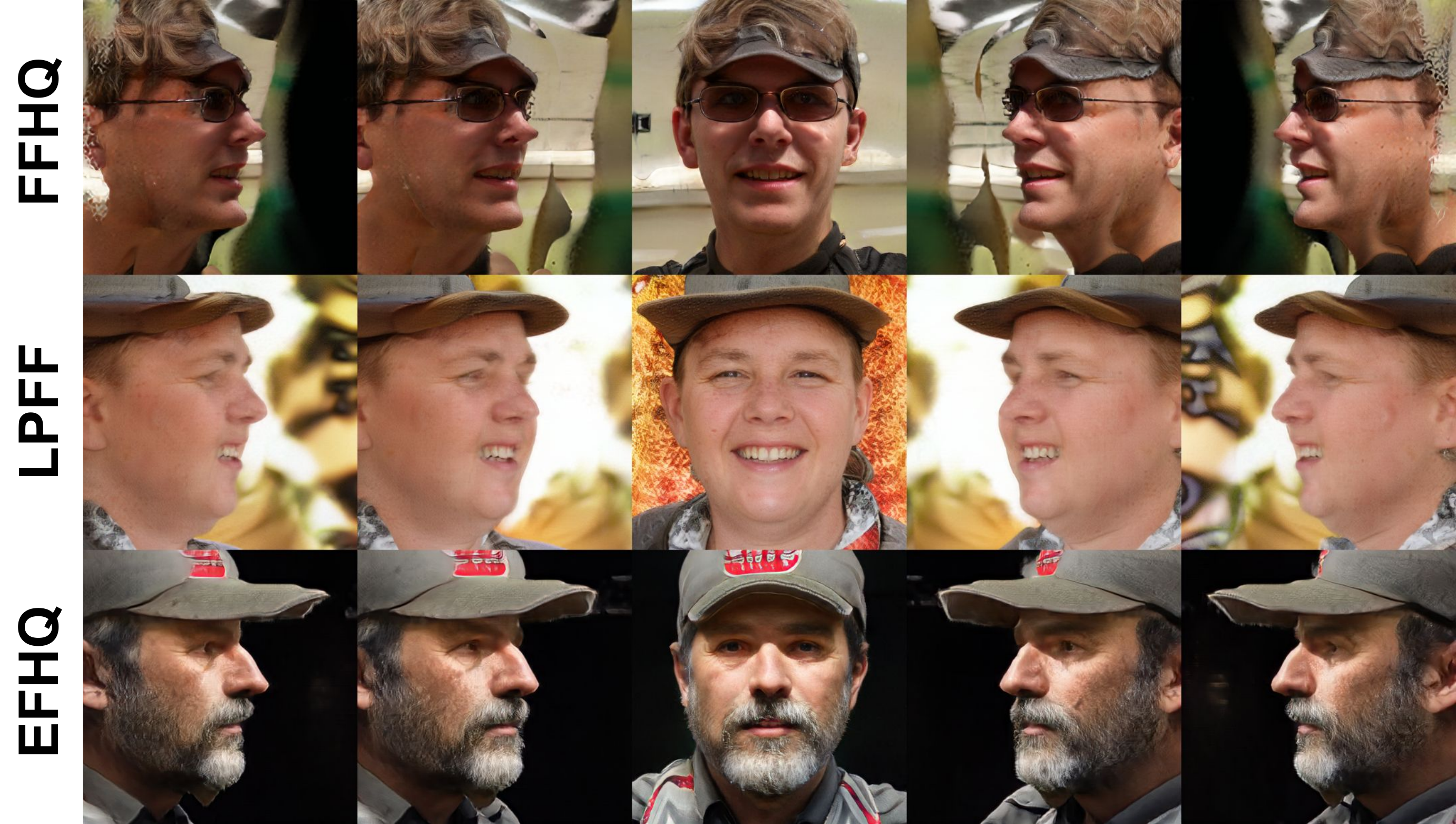}
    \vspace{-2mm}
\caption{\textbf{Comparison} between multiview generated samples, with truncation $\psi=0.8$, of EG3D model trained with various datasets.}
    \vspace{-2mm}
\label{fig:eg3d_compare}
    \vspace{-4mm}
\end{figure}

\subsection{EG3D}
\minisection{Training details.}
    We followed the same pipeline and hyper-parameters as the original EG3D model \cite{eg3d}. However, based on empirical results, we adjusted $\gamma=5$. We also fixed the triplane implementation bug (XY, XZ, ZX) \cite{eg3dbug} to replicate the original idea in the papers (XY, XZ, ZY). We trained the model on 8 Nvidia A100 40GB GPUs for approximately 7 days until convergence.

\minisection{Evaluation Protocol.} 
    To evaluate the models, mirroring \cite{eg3d}, we measure image quality with FID;  multi-view consistency and identity preservation with mean cosine similarity of pre-trained face recognition model\cite{deng2019arcface}; pose accuracy and geometry by calculating MSE against pseudo labels from \cite{deep3dreconstruct}. We evaluate all models on 1,024 generated images, except for FID, which we compute on 50k images.

\minisection{Experimental results.} 
    To evaluate the benefits of our dataset, we trained the same backbone architecture on different datasets: FFHQ ($E^{\text{FFHQ}}$), FFHQ+LPFF ($E^{\text{FFHQ+LPFF}}$), and FFHQ+EFHQ (our model, $E^{\text{FFHQ+EFHQ}}$). The quantitative results are presented in Table \ref{tab:eg3d}, and qualitative examples are shown in Figure \ref{fig:eg3d_compare}. Qualitatively, we observe that $E^{\text{FFHQ+EFHQ}}$ generates improved face shapes in extreme poses compared to $E^{\text{FFHQ}}$, similar to the improvements seen with $E^{\text{FFHQ+LPFF}}$. Quantitatively, the FID of $E^{\text{FFHQ+EFHQ}}$ is comparable to $E^{\text{FFHQ}}$ when evaluated on FFHQ, indicating no degradation in frontal view performance. Meanwhile, the competitive FID achieved by $E^{\text{FFHQ+EFHQ}}$ on the combined FFHQ+EFHQ dataset also suggests proficient modeling of profile views. Regarding identity consistency, we see a significant drop when the reference dataset contains profile-view images, due to the weak confidence of the pretrained face recognition model when handling profile-view images, as confirmed by its low average cosine similarity of around 0.6 on the simple frontal-to-profile verification set CFP-FP\cite{cfpw}. Furthermore, our model also outperforms in pose accuracy and geometry quality with both frontal and cross-pose dataset.

\begin{table}
  \centering
  \begin{tabular}{@{}llcc@{}}
    \toprule
    Metrics & Dataset & Pretrained ControlNet & Ours \\
    \midrule
    NME$\downarrow$ & LPFF & 0.085 & \textbf{0.084} \\
    FID$\downarrow$ & LPFF & 119.74 & \textbf{83.39} \\
    $\text{FID}_{eyepatch}\downarrow$ & LPFF & 103.77 & \textbf{49.45} \\
    \midrule
    NME$\downarrow$ & EFHQ & 0.106 & \textbf{0.081} \\
    FID$\downarrow$ & EFHQ & 111.23 & \textbf{81.90} \\
    $\text{FID}_{eyepatch}\downarrow$ & EFHQ & 137.43 & \textbf{58.60} \\
    \bottomrule
  \end{tabular}
    \vspace{-2mm}
  \caption{\textbf{Quantitative results} of fine-tuning StableDiffusion v1.5 with ControlNet trained on various datasets, using Normalized Mean Error (NME), FID, and FID on eye region.}
  \label{tab:controlnet}
\vspace{-4mm}
\end{table}


\begin{figure}[h]
\centering
\begin{subfigure}{0.45\textwidth}
    \includegraphics[width=\linewidth]{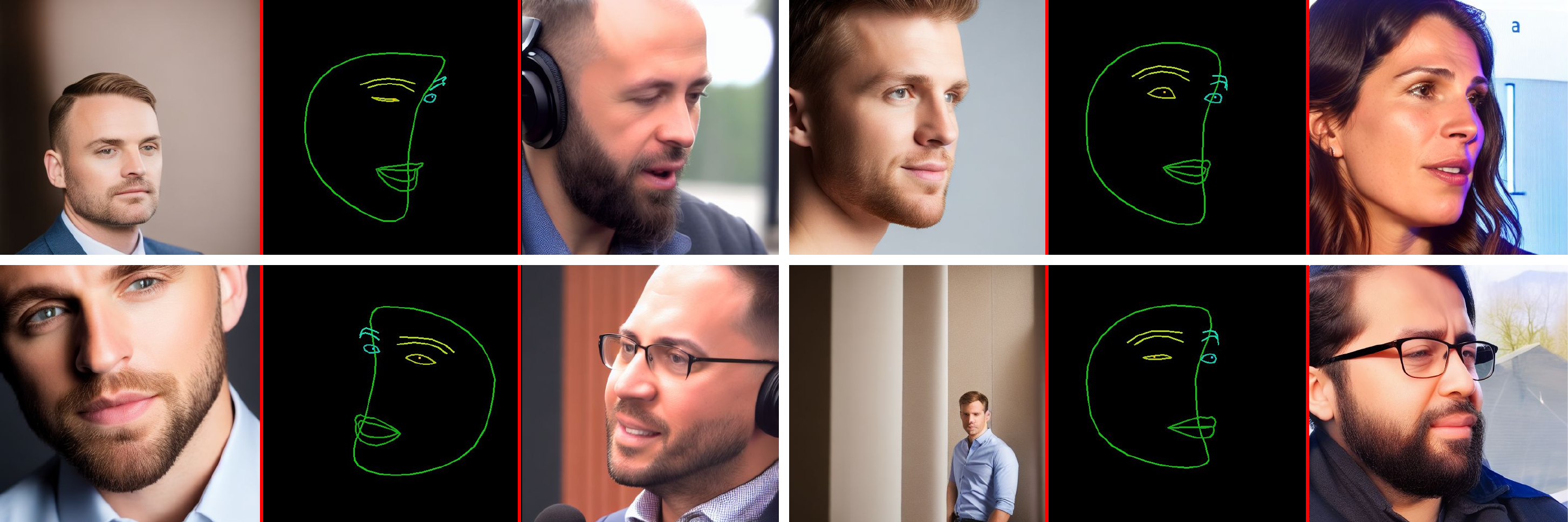}
    \vspace{-4mm}
    \caption{Generated samples; condition image is in between samples.}
    \label{fig:controlnet_compared}
\end{subfigure}
~
\centering
\begin{subfigure}{0.48\textwidth}
    \includegraphics[width=\linewidth]{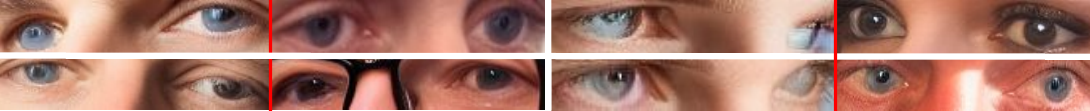}
    \vspace{-4mm}
    \caption{Generated eye regions.}
    \label{fig:controlnet_eyepatch}
\end{subfigure}
    \vspace{-6mm}
\caption{\textbf{Comparison} profile-view generated samples of pretrained ControlNet (left) and our fine-tuned ControlNet (right) with the prompt: ``A profile portrait image of a person".}
    \vspace{-4mm}
\end{figure}

\subsection{ControlNet}
\minisection{Training details.}
    We finetuned Stable Diffusion v1.5 \cite{stablediffusion} using ControlNet \cite{controlnet} on the EFHQ dataset to enhance its ability to generate extreme facial poses. We use AdamW optimizer \cite{adamw} with learning rate $1\times e^{-5}, \beta_0=0.9, \beta_1=0.999$, and $\epsilon=0.01$. We train with batch size 24 on 1 Nvidia A100 40GB GPU in 2 days.

\minisection{Evaluation Protocol.} To assess image qualities across diverse poses, we measure the Normalized Mean Error (NME) of 478-point facial landmarks between the generated and source image. Specifically, we evaluate performance across the full pose distribution in the reference dataset. For the prompt, we use: ``A profile portrait image of a person." with extra positive and negative keywords, further detailed in supplementary.
Utilizing LPFF and EFHQ datasets as the reference datasets, we compute FID to assess synthesized face quality and extend evaluation to cropped eye regions due to unrealistic gazes observed from the original ControlNet model.
Note that the models receive condition images created akin to its training data.

\minisection{Experimental results.} 
\cref{fig:controlnet_compared} visually compares the baseline ControlNet OpenPose face and our fine-tuned model. Our fine-tuned model adeptly conditions diverse extreme poses, generating high-quality results with precise details while maintaining fidelity to the condition, such as mouth shape, and exhibiting minimal artifacts. These patterns indicate the superiority of the fine-tuned model over the baseline model in handling extreme viewing angles. Additionally, \cref{fig:controlnet_eyepatch} contrasts the eye regions of both models. Our fine-tuned model accurately captures natural eye gaze and shape from condition images, a notable improvement over the baseline. Quantitative results in \cref{tab:controlnet} consistently favor our fine-tuned model across all metrics. FID scores on full and eye patches exhibit significant enhancement, and NME performance remains stable on training and reference datasets (EFHQ and LPFF). These findings showcase the fine-tuned model's capability to generate high-quality, geometrically consistent faces across diverse poses.
\section{Face Reenactment Subtask}
\label{sec:facereenact}
\begin{table}
  \centering
  \resizebox{\columnwidth}{!}{
    \begin{tabular}{lcc}
      \hline
      \multirow{2}{*}{Model} & \multicolumn{2}{c}{Test Dataset ($\mathcal{L}_1$ $\downarrow$/ \red{AKD} $\downarrow$/ \blue{AED} $\downarrow$)}  \\
      \cline{2-3}
                               & VoxCeleb1                                    & EFHQ (Ours)       \\
      \hline
      $TPS^{\text{VoxCeleb1}}$ & \textbf{0.041}/\red{\textbf{1.214}}/\blue{\textbf{0.132}} & 0.061/\red{3.814}/\blue{0.275} \\
      $TPS^{\text{VoxCeleb1+EFHQ}}$ &
      \textbf{0.041}/\red{1.250}/\blue{0.133}     & \textbf{0.051}/\red{\textbf{1.963}}/\blue{\textbf{0.214}}\\
      \hline
      $LIA^{\text{VoxCeleb1}}$ & \textbf{0.044}/\red{\textbf{1.342}}/\blue{\textbf{0.138}} & 0.059/\red{3.807}/\blue{0.266} \\
      $LIA^{\text{VoxCeleb1+EFHQ}}$                 &
      \textbf{0.044}/\red{1.398}/\blue{0.142} & \textbf{0.054}/\red{\textbf{2.291}}/\blue{\textbf{0.233}} \\
      \hline
    \end{tabular}}
  \vspace{-2mm}
  \caption{\textbf{Quantitative results} of two state-of-the-art reenactment methods (TPS and LIA) when training on VoxCeleb1 and the combined set VoxCeleb1+EFHQ. We present results evaluated either on VoxCeleb1 (left) or EFHQ (right) test sets.}
  \vspace{-2mm}
  \label{tab:reenactment}
  \vspace{-4mm}
\end{table}
\begin{figure*}[ht!]
\centering
\includegraphics[width=0.8\linewidth]{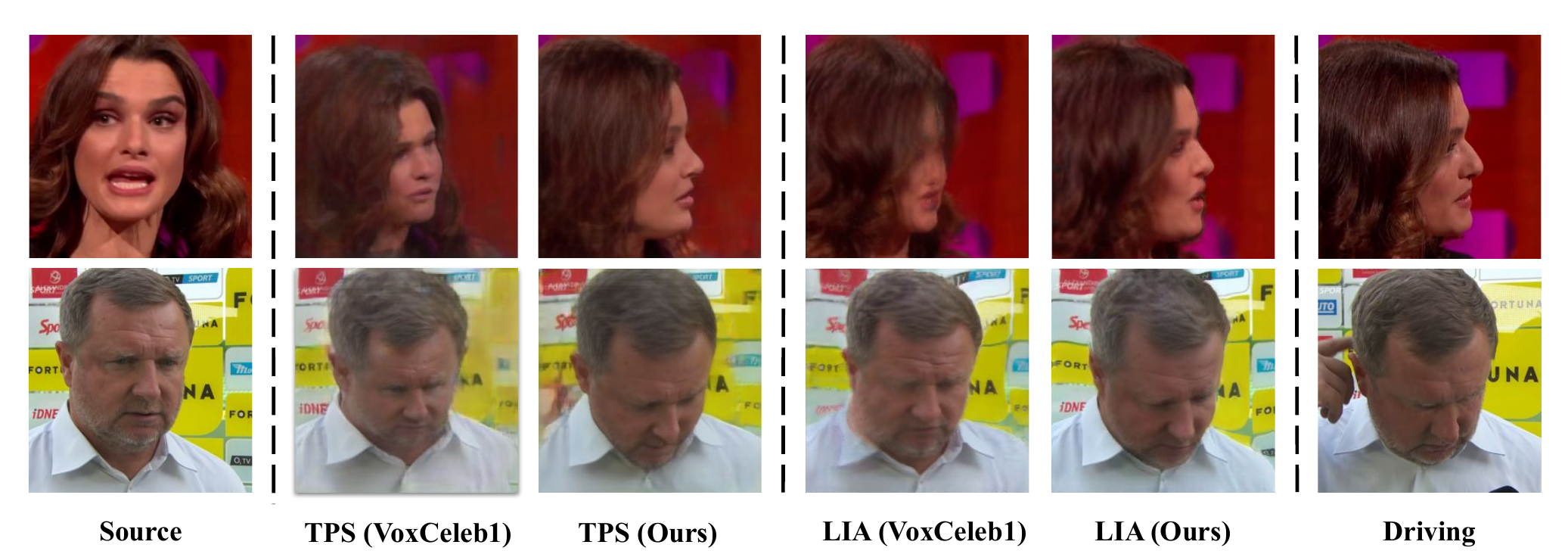}
    \vspace{-2mm}
\caption{\textbf{Comparison} of performance for frontal-profile face reenactment on EFHQ. From left to right: source image, TPS on VoxCeleb1, TPS with EFHQ, LIA on VoxCeleb1, LIA with EFHQ, driving image. Additional EFHQ training data improves synthesis of extreme poses.}
\label{fig:reenact_compared}
    \vspace{-4mm}
\end{figure*}




\subsection{Evaluation Protocol} 
We follow the same preprocessing and evaluation protocol as FOMM \cite{fomm}. Specifically, all the frames are cropped to $256\times256$, and we use the first frame of a clip as the source image to reconstruct the remaining frames. In the case of the EFHQ test set, most source images have a frontal pose, while the target frames have an extreme pose. Some test cases only contain extreme pose to evaluate self-reconstruction on these cases. We run the experiments using two recent methods, Thin-Plate Spline Motion Model (TPS) \cite{tps} and Latent Image Animator (LIA) \cite{lia}. For training details, please refer to the supplementary. We select the following three metrics to compare the generated frames and the driving ground truth. $\mathcal{L}_1$ measures the average distance among pixel values, indicating reconstruction faithfulness. Average Keypoint Distance (AKD) calculates the mean distance between keypoints, thus assessing the pose accuracy. Average Euclidean Distance (AED) measures identity preservation by computing the average $\mathcal{L}_2$ distance between identity embeddings.

\subsection{Experimental results}


We compare TPS and LIA models trained with our supplementary EFHQ dataset ($TPS^{\text{VoxCeleb1+EFHQ}}$ and $LIA^{\text{VoxCeleb1+EFHQ}}$) against models trained solely on VoxCeleb1 ($TPS^{\text{VoxCeleb1}}$ and $LIA^{\text{VoxCeleb1}}$). \cref{tab:reenactment} displays quantitative results, revealing that models trained with EFHQ maintain comparable performance when evaluated on VoxCeleb1, emphasizing high-quality performance in frontal-to-frontal reenactment scenarios. Notably, on the EFHQ test set, both EFHQ-trained models show significant improvement in all metrics, indicating enhanced extreme pose and identity reenactment performance. In \cref{fig:reenact_compared}, qualitative results prove that EFHQ-trained models excel in transferring facial expressions and motion across diverse poses. Models trained on EFHQ showcase improved image quality, reduced artifacts, and well-preserved shapes of driving faces. These results highlight our models' effectiveness in capturing and reproducing facial motions across varying poses while enhancing fidelity to the original facial features.
\section{Face Recognition Subtask}
\label{sec:faceveri}
\subsection{Evaluation Protocol}
    To evaluate face recognition performance on extreme poses, we selected two state-of-the-art methods: ArcFace\cite{deng2019arcface} and AdaFace\cite{kim2022adaface}. We further explored various backbone sizes (ResNet18, ResNet50, ResNet101)\cite{resnet} and training datasets (MS1MV2\cite{ms1m}, Glint360K\cite{glint360k}, WebFace4M\cite{webface260m}) of increasing size. For metrics, we used True Acceptance Rate (TAR) at False Acceptance Rate (FAR) of $1\times e^{-3}$.

\begin{table}
  \centering
  \resizebox{\columnwidth}{!}{
  \begin{tabular}{@{}llcccc@{}}
    \toprule
    \multirow{2.5}{*}{Model}& \multirow{2.5}{*}{Trainset} & \multirow{2.5}{*}{CPLFW}&  \multicolumn{3}{c}{EFHQ} \\
         \cmidrule(lr){4-6}
     &  &  & f2f & f2p & p2p \\
    \midrule
    $R18_{\text{arcface}}$ & Glint360k & 0.80 & 0.94 & $\boldsymbol{0.57_{\red{0.37}}}$ & $0.88_{\red{0.06}}$ \\
    $R18_{\text{arcface}}$ & MS1MV3 & 0.75 & 0.93 & $\boldsymbol{0.57_{\red{0.36}}}$ & $0.88_{\red{0.05}}$ \\
    $R50_{\text{arcface}}$ & Glint360k & 0.91 & 0.97 & $\boldsymbol{0.77_{\red{0.20}}}$ & $0.92_{\red{0.05}}$ \\
    $R100_{\text{arcface}}$ & Glint360k & 0.93 & 0.98 & $0.83_{\red{0.15}}$ & $\boldsymbol{0.82_{\red{0.16}}}$ \\
    $R100_{\text{adaface}}$ & WebFace4M & 0.94 & 0.99 & $\boldsymbol{0.94_{\red{0.05}}}$ & $0.97_{\red{0.02}}$ \\
    \bottomrule
  \end{tabular}
  }
    \vspace{-2mm}
  \caption{\textbf{Quantitative evaluation} of state-of-the-art face recognition models on the cleansed CPLFW and the proposed benchmark dataset. We use the metric TAR@FAR=1e-3 (higher is better). Our EFHQ benchmark includes frontal-frontal (f2f), frontal-profile (f2p), and profile-profile (p2p) subsets. For f2p and p2p, we report the score drop when comparing to f2f in the \red{red} subscription. The lowest benchmark score for each model is in \textbf{bold}.}
  \label{tab:facevec}
\vspace{-6mm}
\end{table}


\subsection{Experimetal results}

Before assessing face recognition networks on EFHQ, we first clean the commonly-used version of CPLFW, processed by\cite{deng2019arcface}, by removing misaligned images (please refer to the Supplementary), and evaluate the models on it. 
Models performed well on this cleansed set, with minimal gaps between ArcFace and AdaFace.

Our proposed evaluation benchmark involves three pose-based scenarios. Results in \cref{tab:facevec} reveal performance drops of 5-37\% on the frontal-profile subset compared to frontal-frontal, highlighting the challenge of matching faces in different poses. Larger ResNet models outperformed ResNet18 across subsets, showcasing the advantages of increased capacity in handling pose variation. However, the ResNet100 model trained on Glint360K performed worse than smaller models on the profile-profile subset, indicating the need for sufficient profile data to prevent overfitting. Our results also indicate AdaFace's suitability for handling pose variations. 
In summary, pose mismatch brings significant challenges for face recognition, necessitating larger models trained on diverse data covering profile faces.
\section{Discussion}
\label{sec:discussion}
\begin{figure}[ht!]
\vspace{-3mm}
\centering
\includegraphics[width=0.9\linewidth]{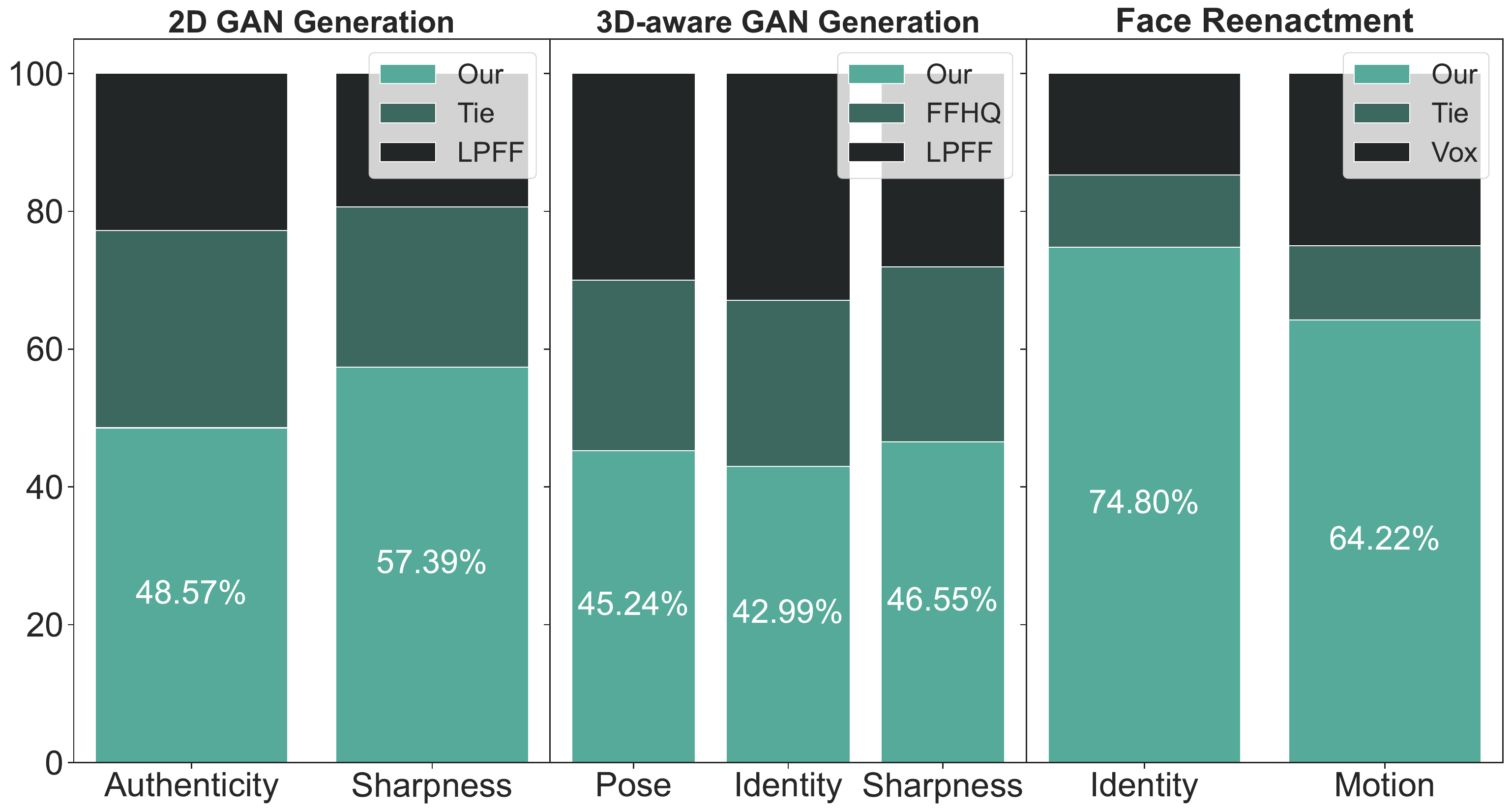}
    \vspace{-2mm}
\caption{\textbf{User study results.} We reported the percentage of times testers rank whose output is best or tied for each criterion.}
\label{fig:survey_all}
\end{figure}

\minisection{User survey.} We conduct user surveys to evaluate the face generation improvement of our method through human perceptual assessment. In this survey, we compare the outputs of the models mentioned in \cref{sec:facegen,sec:facereenact}. The setup of this user survey is included in the supplementary. In all human evaluation results (\cref{fig:survey_all}), our models outperform previous ones, notably on identity and motion test of face reenactment models (74.80\% and 64.22\%, respectively).

\begin{figure}[ht!]
\centering
\includegraphics[keepaspectratio,width=0.85\linewidth]{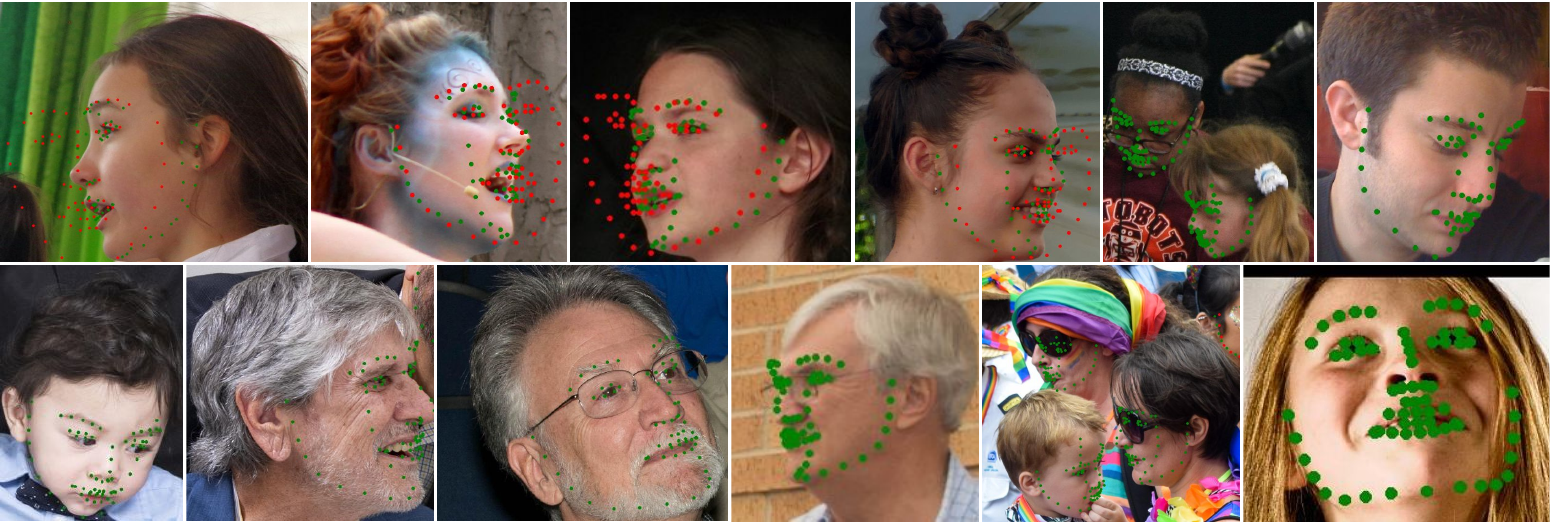}
    \vspace{-2mm}
\caption{\textbf{Representative examples} of the raw LPFF dataset that got excluded from the final dataset, either from misdetections or pose filtering process. \textcolor{red}{\textbf{Red}} keypoints represents landmarks prediction of selected samples from LPFF, \applegreen{\textbf{green}} keypoints represents landmarks prediction from our pipeline.}
\label{fig:pipeline_robustness}
\vspace{-4mm}
\end{figure}

\minisection{Robustness of data processing pipeline.} 
As shown in \cref{fig:pipeline_robustness}, our data processing pipeline can detect additional profile-view faces filtered out by the LPFF pipeline, likely due to misdetections or incorrect pose filtering. This confirms the robustness of our pipeline, and we hope it will enable the community to develop larger and higher-quality in-the-wild facial datasets.

\section{Conclusion}
\label{sec:conclusion}

    This work has introduced a large-scale, diverse facial dataset to address performance gaps between frontal and profile faces, powered by a novel and robust data processing pipeline. Crucially, we provided tailored sub-datasets for advancing essential tasks like face synthesis and reenactment. 
    Moreover, our new face verification benchmark reveals gaps between techniques, granting valuable insights. Ultimately, we hope this high-quality, diverse facial dataset opens up new and exciting opportunities to push forward cross-pose tasks.
{
    \small
    \bibliographystyle{ieeenat_fullname}
    \bibliography{main}
}

\clearpage
\setcounter{page}{1}
\maketitlesupplementary

\section{More on dataset pipeline}
\subsection{Pose bin}
\begin{figure}[ht!]
\centering
\includegraphics[width=0.95\linewidth]{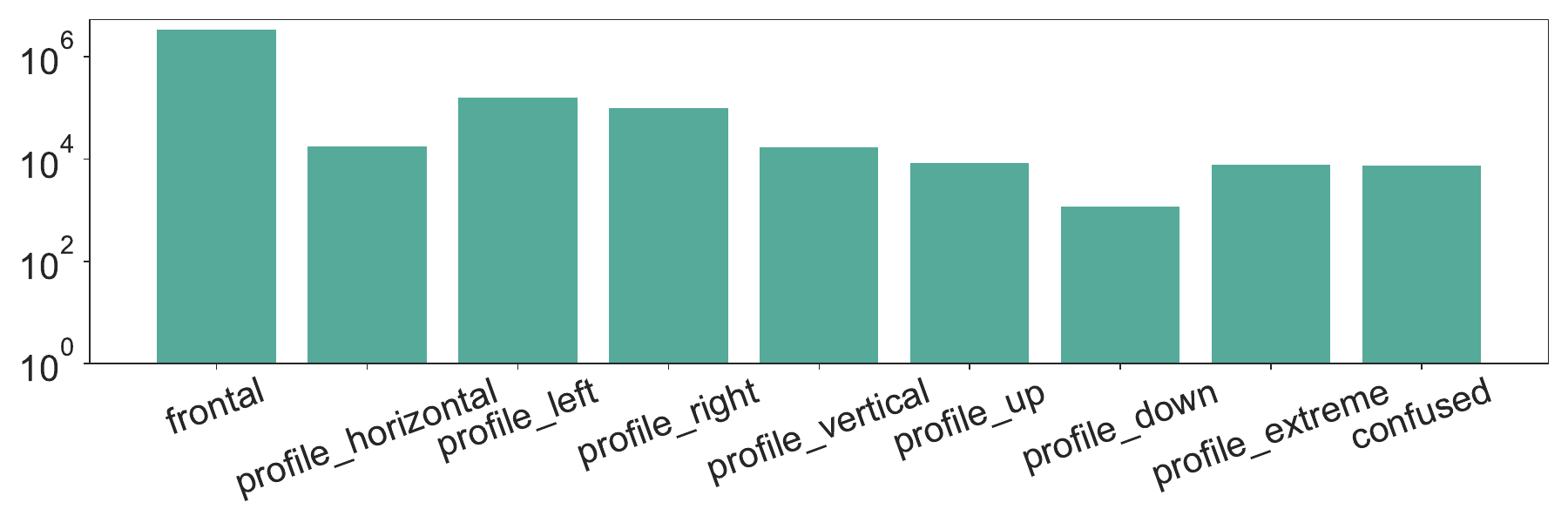}
    \vspace{-3mm}
\caption{\textbf{Frame distribution} for each pose bin after our automated attributes preparation process.}
\label{fig:bin_dist}
\vspace{-3mm}
\end{figure}

\begin{figure}[ht!]
\centering
\includegraphics[width=0.95\linewidth]{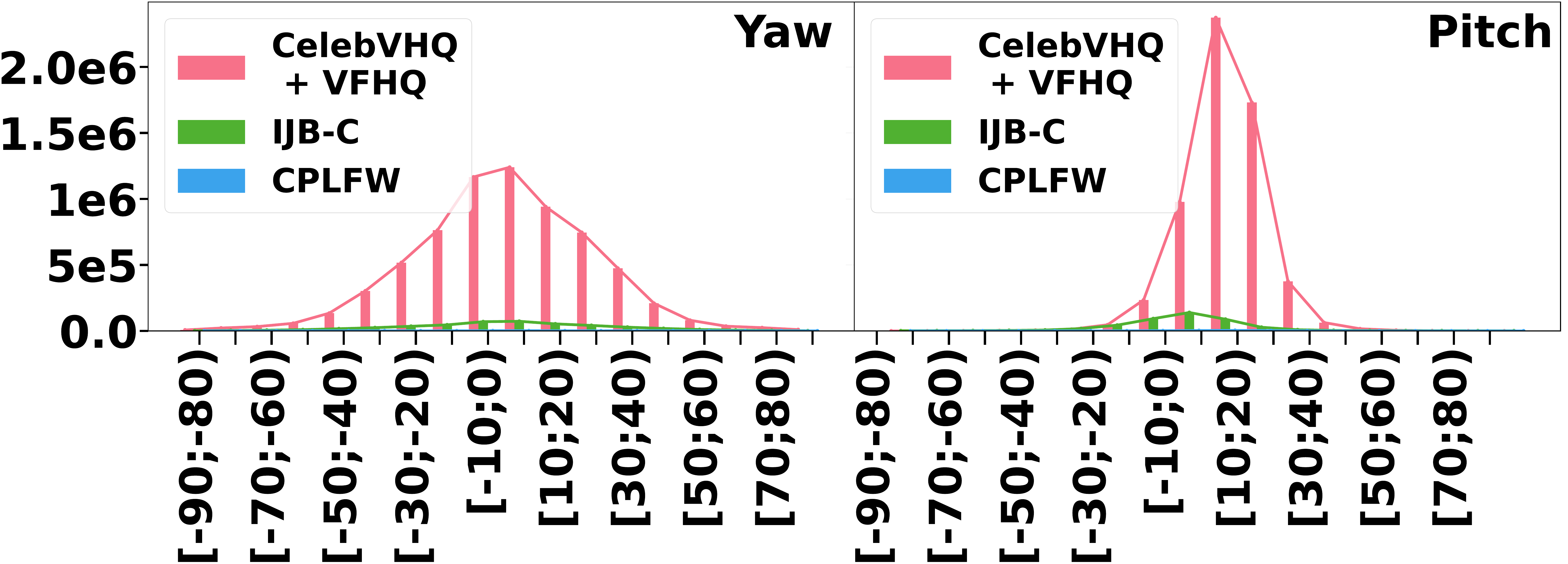}
    \vspace{-3mm}
\caption{\textbf{Pose distribution} in terms of angle bins of CelebV-HQ + VFHQ, IJB-C, CPLFW}
\label{fig:supp_wholedist}
\vspace{-3mm}
\end{figure}

\cref{fig:bin_dist} illustrates the distribution of frames per bin following our automated processing pipelines, presented on a logarithmic scale. As depicted in the figures, the original datasets contain a substantial volume of frames, yet a predominant portion consists of frontal faces. This observation substantiates our previously articulated arguments regarding the prevalent bias toward frontal views. Leveraging our preprocessing pipelines enables us to concentrate solely on extreme view images, reserving the utilization of frontal views for some specific subtasks like face reenactment and face verification. Furthermore, \cref{fig:supp_wholedist} shows the pose distribution in terms of angle bins of the original two video-dataset compared to IJB-C and CPLFW.

\subsection{Model hyperparameters} In the process of extracting face bounding boxes and 5 keypoints via RetinaFace\cite{retinaface}, we keep only detected faces with confidence above 0.9 to ensure the retention of profile faces while minimizing false negatives and low quality facial detections. Additionally, faces with an area smaller than $256 \times 256$ in both datasets are disregarded. To further refine the selection, the HyperIQA score, specifically set at 42 according to \cite{vfhq}, is employed during the sampling phase to exclude frames of poor quality. Moreover, to address the issue of low-light images, which is not explicitly assessed by HyperIQA, a method referenced in \cite{brightness} is utilized, filtering out images with a brightness value below 40, a value determined empirically to suit our specific requirements. For DirectMHP, we first apply multiscale inference and later use a confidence threshold of 0.7 and an IoU threshold of 0.45 during the non max suppression (NMS) algorithm. All other models, if not specified, follow the original setup of the original authors.

\section{Dicussion on Negative Socieltal Impacts}
\label{sec:supp_negative_impact}
Our dataset comes from putting together two public datasets \cite{vfhq, celebvhq}. Consequently, we share similar concerns regarding privacy and bias with the original datasets from which our data is sourced. It is crucial for any user of our dataset to adhere strictly to the rules outlined in the original datasets regarding its usage. Moreover, we advocate and encourage further exploration and endeavor aimed at developing strategies that proactively minimize and counteract any deleterious societal implications, thereby harnessing the substantial research value embedded within the dataset while upholding ethical considerations and societal well-being.

\section{More on face recognition task}
\subsection{Face identification}
Besides face verification, our dataset can also be used for face identification evaluation. Since we have multiple frames and multiple videos for an identity, we decided to develop both open-set and closed-set scenarios for the new benchmarking dataset. In the closed-set, we used 10,000 gallery items (1 per ID) and 50,000 probes. For open-set, 1,000 gallery items (1 per ID) and 5,000 mate searches, plus 32,275 non-mate searches from other IDs. For filtering and augmentation, we apply the same strategy as when building the face verification benchmark dataset. We followed NIST guidelines for metrics: FNIR at FPIR for open-set and TPIR at Ranks for closed-set. We tested Arcface Resnet-18/50 pretrained on Glint360K, \cref{tab:faceid_compare} shows that pose remains a challenge in the face identification task.

\begin{table}
\begin{tabular}{ccccccc}
        \hline
        & \multicolumn{3}{c}{\textbf{Open-set}} & \multicolumn{3}{c}{\textbf{Closed-set}} \\ \hline
        & \textbf{FPIR} & 0.003 & 0.03 & \textbf{Rank} & 1 & 5 \\ \hline
        $\text{R18}$ & \textbf{FNIR} $\downarrow$ & 0.45 & 0.26 & \textbf{TPIR} $\uparrow$ & 0.80 & 0.96 \\
        $\text{R50}$ & \textbf{FNIR} $\downarrow$ & 0.41 & 0.24 & \textbf{TPIR} $\uparrow$ & 0.83 & 0.97 \\ \hline
\end{tabular}
\caption{\textbf{Quantitative result} of state-of-the-art models upon our proposed face identification benchmark dataset. We provided both scenarios: open-set and closed-set.}
\label{tab:faceid_compare}

\end{table}

\begin{table}
\centering
\resizebox{.95\columnwidth}{!}{%
\begin{tabular}{lllllll}
    \hline
     \multirow{2}{*}{Training} &
      \multicolumn{3}{c}{EFHQ} &
      \multicolumn{1}{c}{\multirow{2}{*}{\begin{tabular}[c]{@{}c@{}}CP\\ LFW\end{tabular}}} &
      \multicolumn{1}{c}{\multirow{2}{*}{\begin{tabular}[c]{@{}c@{}}CF\\ PFP\end{tabular}}} &
  \multicolumn{1}{r}{\multirow{2}{*}{IJBC}} \\ \cline{2-4}
          & \multicolumn{1}{c}{f2f} & \multicolumn{1}{c}{f2p} & \multicolumn{1}{c}{p2p} & \multicolumn{1}{c}{} & \multicolumn{1}{c}{} & \multicolumn{1}{r}{} \\ \hline
    MS1MV3 & 0.93 & 0.57 & 0.88 & 0.75                 & 0.82                 & 0.94                 \\
    + EFHQ      & $\boldsymbol{0.98}$ & $\boldsymbol{0.77}$ & $\boldsymbol{0.94}$ & $\boldsymbol{0.79}$                 & $\boldsymbol{0.90}$                 & 0.94                 \\ \hline
\end{tabular}
}
\caption{Effect of \textbf{EFHQ as additional training data} on model performance, with TAR@FAR=1e-4 for IJB-C and TAR@FAR=1e-3 for others.
\label{tab:supp_finetune}}

\end{table}

\subsection{Training face recognizer}
For futher investigation of the pose distribution, we developed a training dataset of around 300,000 samples taken from identities that were not used for the proposed benchmark dataset. Due to time limitations, we trained only a ResNet-18 model using MS1MV3 along with this dataset for 20 epochs, following the same training settings from \cite{deng2019arcface}. As shown if \cref{tab:supp_finetune}, our training dataset enhances the trained model's performance on pose-focused benchmarks, albeit not marginally significant.

\section{Training details}
\label{sec:supp_trainingdetails}
\subsection{Face Reenactment}
For face reenactment, we retrained two current state-of-the-art models, TPS and LIA, starting from scratch. This retraining was conducted on two datasets, including VoxCeleb1 and VoxCeleb1 supplemented with our EFHQ dataset. All models were trained using 2 Nvidia A100 GPUs with 40GB of memory. For both TPS and LIA models, we followed the hyperparameter choices outlined in the respective papers \cite{tps, lia}.

In the case of VoxCeleb1, we followed the download and preprocessing pipeline as described in \cite{fomm}. However, due to copyright and regional restrictions, we encountered limitations in fully acquiring the original dataset. Specifically, out of the 3,442 videos available in the original data, we were only able to download approximately 3,000 videos. This constraint limited our capacity to match the performance of the released pretrained models. Consequently, we opted to train both models from scratch for both datasets to ensure a fair benchmark for evaluation.

\minisection{Thin-Plate Spline Motion Model}.
We set the total batch size to 28. The training process consists of two phases: the Base model phase, spanning 100 epochs with a learning rate of 0.0002, followed by the AVD Network phase, which extends for 200 epochs with a learning rate of 0.001. We utilize $K=10$ TPS transformations to approximate the motion. In our loss function, we incorporate $\lambda=10$ for the perceptual loss.

\minisection{LIA}.
For LIA, we train both models with a total batch size of 32 and a learning rate of 0.002. The dimensions of all latent codes and the directions in the set of motion directions $D_{m}$ are set to be 512. Meanwhile, we set the number of motion directions $D_{m}$ to 20. In our loss function, we use $\lambda=10$ for the perceptual loss.

\subsection{ControlNet}
Throughout both the training and sampling stages, a standardized set of prompt templates was consistently employed across all samples, formulated as ``\textit{A profile portrait image of a [emotion] [race] [gender].}" Specifically, in the sampling phase for evaluation, we implemented an augmentation strategy by integrating positive attributions, encompassing descriptors like ``\textit{rim lighting, studio lighting, dslr, ultra quality, sharp focus, tack sharp, depth of field (dof), film grain, Fujifilm XT3, crystal clear, 8K UHD, highly detailed glossy eyes, high detailed skin, and skin pores.}" Complementing these positive descriptors, we incorporated a set of negative keywords, including ``\textit{disfigured, ugly, bad, immature, cartoon, anime, 3d, painting, and black and white.}" The intent behind this approach was to refine the image quality within the sampling and evaluation process. This strategic augmentation framework aimed to enhance the overall quality and fidelity of generated images, ensuring a more refined output aligning with desired criteria and minimizing undesirable attributes.

\section{CPLFW artifacts}
\begin{figure}[ht!]
\centering
\includegraphics[keepaspectratio,width=\linewidth]{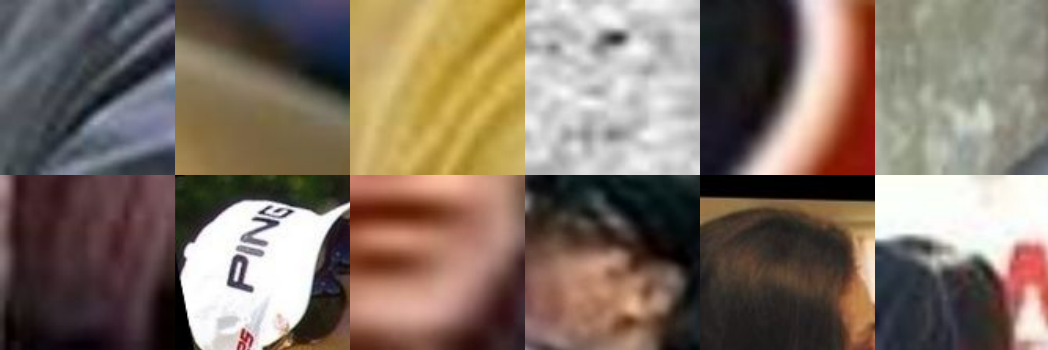}
\caption{Representative examples of the noise sample due to misalignments in the widely-used processed version of CPLFW.}
\label{fig:cplfw_bad}
\end{figure}
    During the examination of the commonly utilized processed version of CPLFW\cite{cplfw} by \cite{deng2019arcface}, numerous instances of noise were identified, encompassing non-human and partial-human images. These anomalies likely arose from misalignments attributed to flawed landmark detections. To systematically address the bulk of these artifacts, we integrated a pretrained RetinaFace model \cite{retinaface} to automatically filter out images lacking detections. This filtering process resulted in the removal of 696 images, equating to 696 pairs, constituting approximately 11\% of the total pairs. \cref{fig:cplfw_bad} showcases representative samples obtained through our cleaning pipeline. In the first row, we compile non-human images, whereas in the second row, we compile partial-human images.
    
\section{More on the robustness of data processing pipeline.} 
To evaluate the robustness of our data processing pipeline, we test it with the raw LPFF dataset. As shown in \cref{fig:pipeline_robustness}, our pipeline detected additional profile-view faces filtered out by the LPFF pipeline, likely due to misdetections or overly strict pose density gating. The LPFF pipeline relies on three main processes for filtering: landmark prediction with Dlib\cite{dlib}, pose density gating using 3D face reconstruction\cite{deep3dreconstruct}, and manual inspection. The latter two stages depend heavily on accurate landmark predictions in the first step. Compared to Dlib, methods like RetinaFace\cite{retinaface} and SynergyNet\cite{synergynet} used in our pipeline offer improved face and landmark detection, especially for challenging poses. Our pipeline delivered markedly much superior landmark quality. To filter frontal samples, the previous work relied solely on the reconstruction model for pose estimation. Although \cite{deep3dreconstruct} can provide state-of-the-art results, it depends on properly aligned images and may produce unreliable predictions when landmark quality is poor. Our pipeline uses an ensemble of pose estimators based on different methods, providing robustness in cases where a single model fails.
By combining multiple pose prediction techniques, our approach can overcome the limitations of relying on just the output of reconstruction models, especially when image alignment is imperfect or landmarks are of low quality. This demonstrates the greater robustness of our approach to diverse poses and image variability compared to prior work. 
Moving forward, we hope our data processing methodology will enable the community to develop larger and higher-quality in-the-wild facial datasets. 
Robust pipelines to handle pose diversity and image noise will be essential to maximize data utilization and quality in these days and age.

\section{Survey setup}
\subsection{2D GAN Generation}
    This survey involves a comparative analysis between samples generated by StyleGAN2-ADA models trained with our setup (FFHQ+EFHQ) versus LPFF setup (FFHQ+LPFF). To ensure fairness, we specifically select pairs of samples exhibiting similar pose angles. Our assessment centers on two primary factors: Authenticity and Sharpness. In evaluating authenticity, participants are tasked with identifying the image with superior human face quality, considering artificial elements like unrealistic human features, skin tone, face shape, and facial accessories. Regarding sharpness, participants are prompted to examine image noisiness and select the image with generally higher image quality. In cases where a clear preference cannot be determined, participants have the option to indicate a tie. A total of 140 participants were engaged, contributing a combined total of 2800 votes over 20 questions.
\subsection{EG3D}
    In this survey, we conduct a comparison between samples generated by EG3D models trained using our setup (FFHQ+EFHQ) and LFPP setup (FFHQ+LPFF). To ensure equitable assessment, we deliberately pair samples with similar gender, expression, and accessories whenever feasible. Each sample is presented through multiple views, including a frontal view for reference and four extreme views for comprehensive evaluation. Our analysis centers on three primary criteria: Pose, Identity, and Sharpness. Regarding pose assessment, participants are tasked with identifying the sample with superior face quality among the extreme views, specifically focusing on facial shape, such as nose and chin, while considering the presence of artifacts like unnatural distortions. In evaluating identity, participants are prompted to gauge how closely the facial identity aligns with the frontal view. Lastly, participants are asked to assess image quality for sharpness, emphasizing less noise and sharper details in the facial region. A total of 145 participants were engaged, contributing a combined total of 2900 votes, over 20 questions, to the study.

\subsection{Face Reenactment}
    This study conducts a comparative analysis of models trained on both VoxCeleb1 and VoxCeleb1+EFHQ datasets for TPS and LIA \cite{tps, lia}. In order to offer a more comprehensive evaluation scenario for users, we compare video output sequences of same-identity reenactment between the baseline model and the model trained on our dataset rather than assessing individual frames. This approach facilitates a nuanced evaluation, especially in capturing the smoothness of pose transitions and illustrating video consistency when handling long sequences of extreme head poses. Our assessment focuses primarily on two key criteria: Identity and Motion. In the context of identity replication, our focus is on evaluating which model more accurately replicates the identity portrayed in the ground-truth driving video. This assessment takes into consideration facial features and the presence of any artificial artifacts. For motion replication, the focus shifts to discerning which model more accurately reproduces the sequence of motion represented in the driving video. In cases where users find it challenging to determine superiority, they have the option to indicate a tie. In total, we involved 170 users, who collectively contributed 8500 votes to this evaluation, addressing 25 questions for each face reenactment method.

\section{Additional Qualitative results}
\begin{figure*}[ht!]
\centering
\includegraphics[keepaspectratio,width=0.9\linewidth]{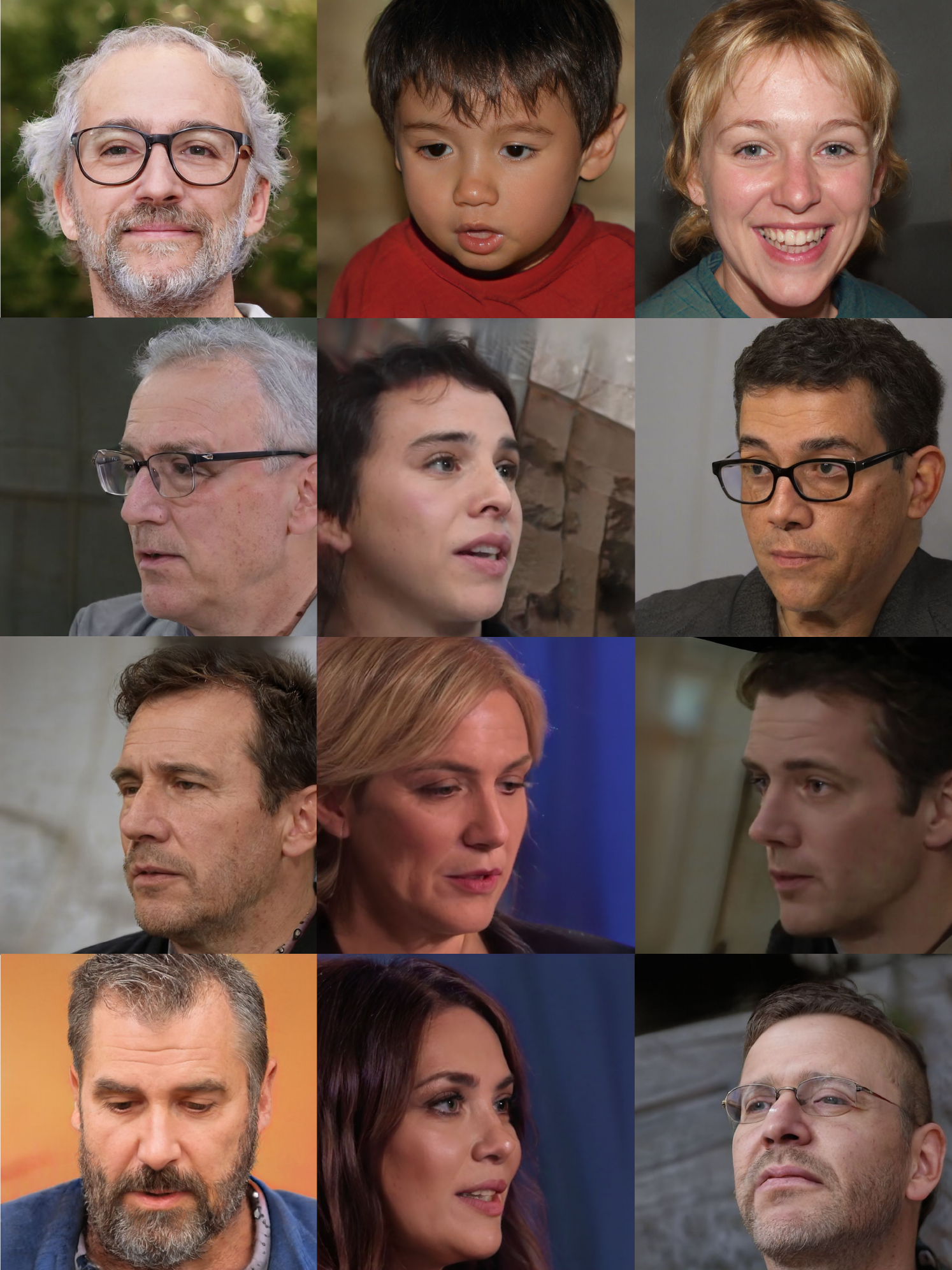}
\caption{Samples from the model trained with FFHQ+EFHQ, with truncation $\psi=0.7$..}
\label{fig:supp_stylegan2}
\end{figure*}

\begin{figure*}[ht!]
\centering
\includegraphics[keepaspectratio,width=0.95\linewidth]{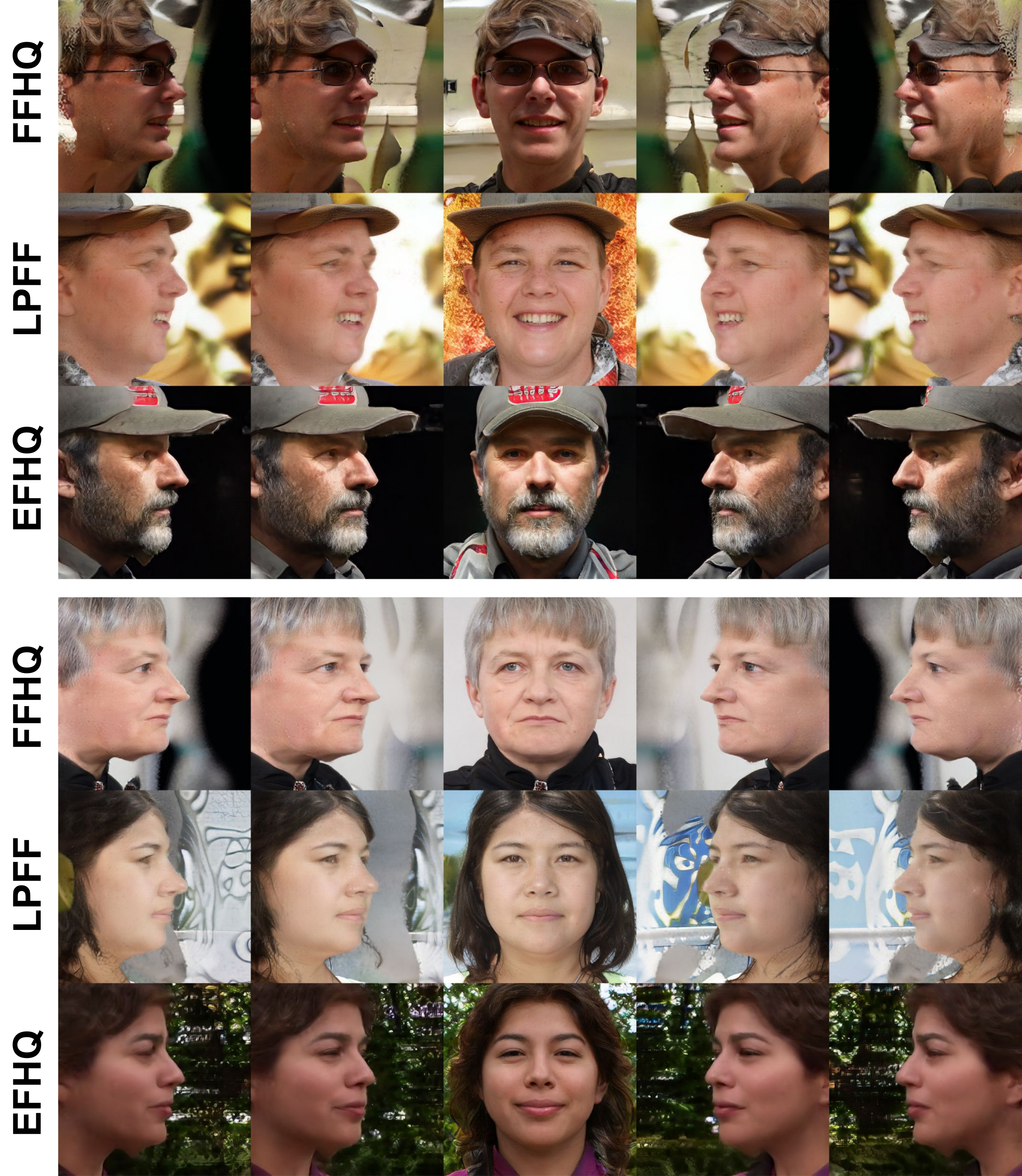}
    \vspace{-2mm}
\caption{\textbf{Comparison} between multiview generated samples, with truncation $\psi=0.8$, of EG3D model trained with various datasets (for both LPFF/EFHQ, the training dataset is combined with FFHQ).}
    \vspace{-2mm}
\label{fig:supp_eg3d1}
    \vspace{-4mm}
\end{figure*}

\begin{figure*}[ht!]
\centering
\includegraphics[keepaspectratio,width=0.95\linewidth]{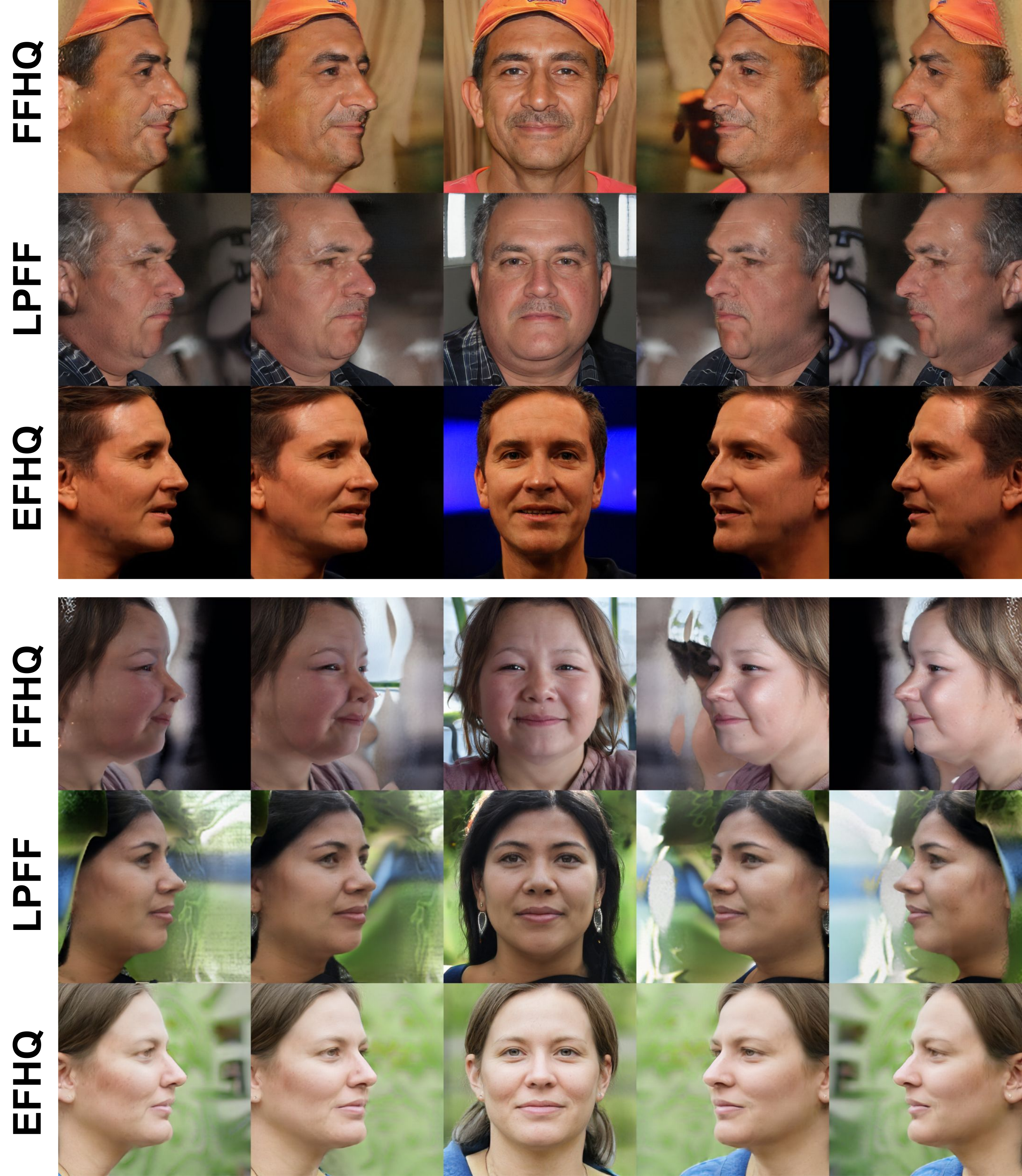}
    \vspace{-2mm}
\caption{\textbf{Comparison} between multiview generated samples, with truncation $\psi=0.8$, of EG3D model trained with various datasets (for both LPFF/EFHQ, the training dataset is combined with FFHQ).}
    \vspace{-2mm}
\label{fig:supp_eg3d2}
    \vspace{-4mm}
\end{figure*}

\begin{figure*}[ht!]
\centering
\includegraphics[keepaspectratio,width=0.95\linewidth]{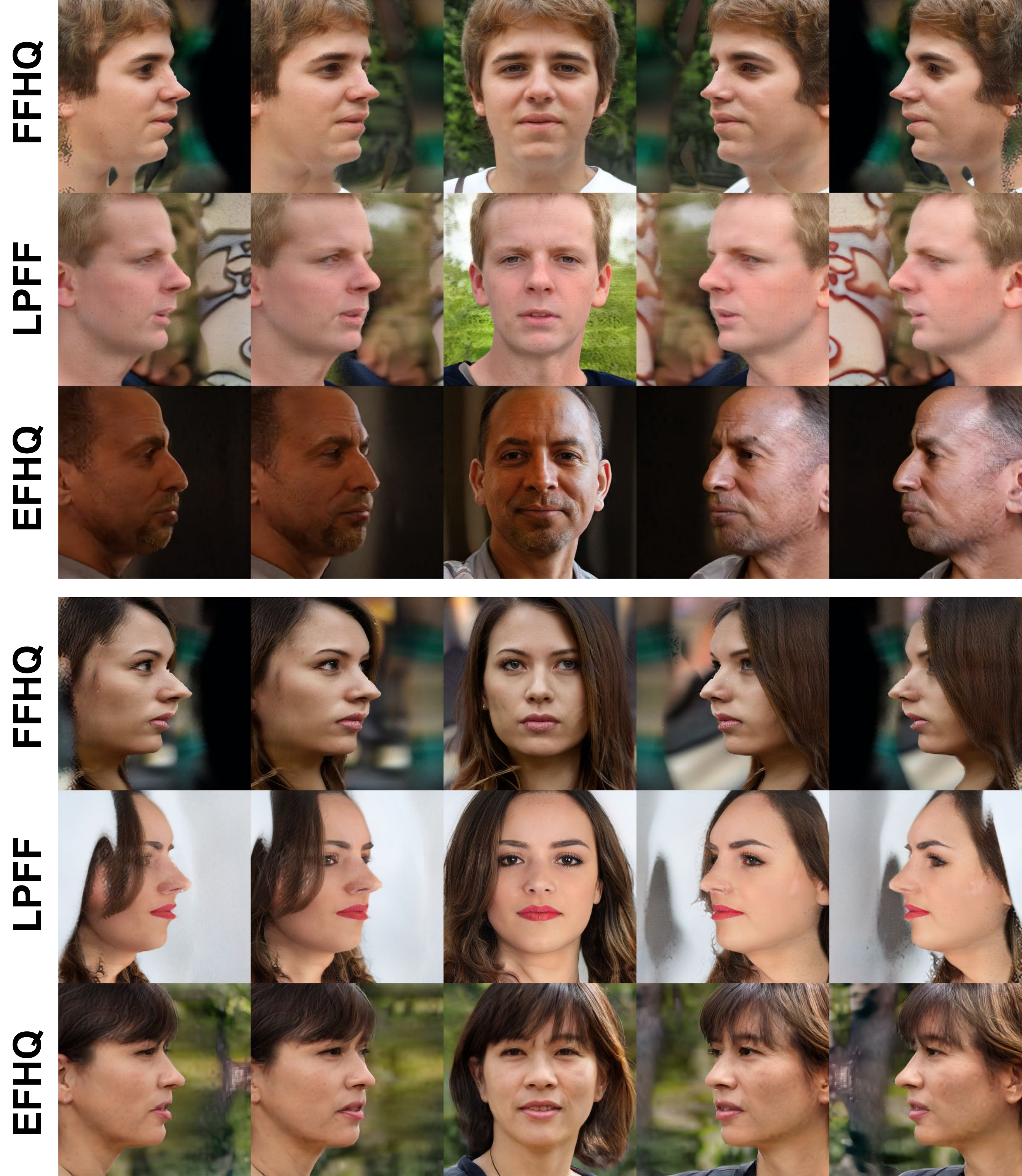}
    \vspace{-2mm}
\caption{\textbf{Comparison} between multiview generated samples, with truncation $\psi=0.8$, of EG3D model trained with various datasets (for both LPFF/EFHQ, the training dataset is combined with FFHQ).}
    \vspace{-2mm}
\label{fig:supp_eg3d3}
    \vspace{-4mm}
\end{figure*}

\begin{figure*}[ht!]
\centering
\includegraphics[keepaspectratio,width=0.95\linewidth]{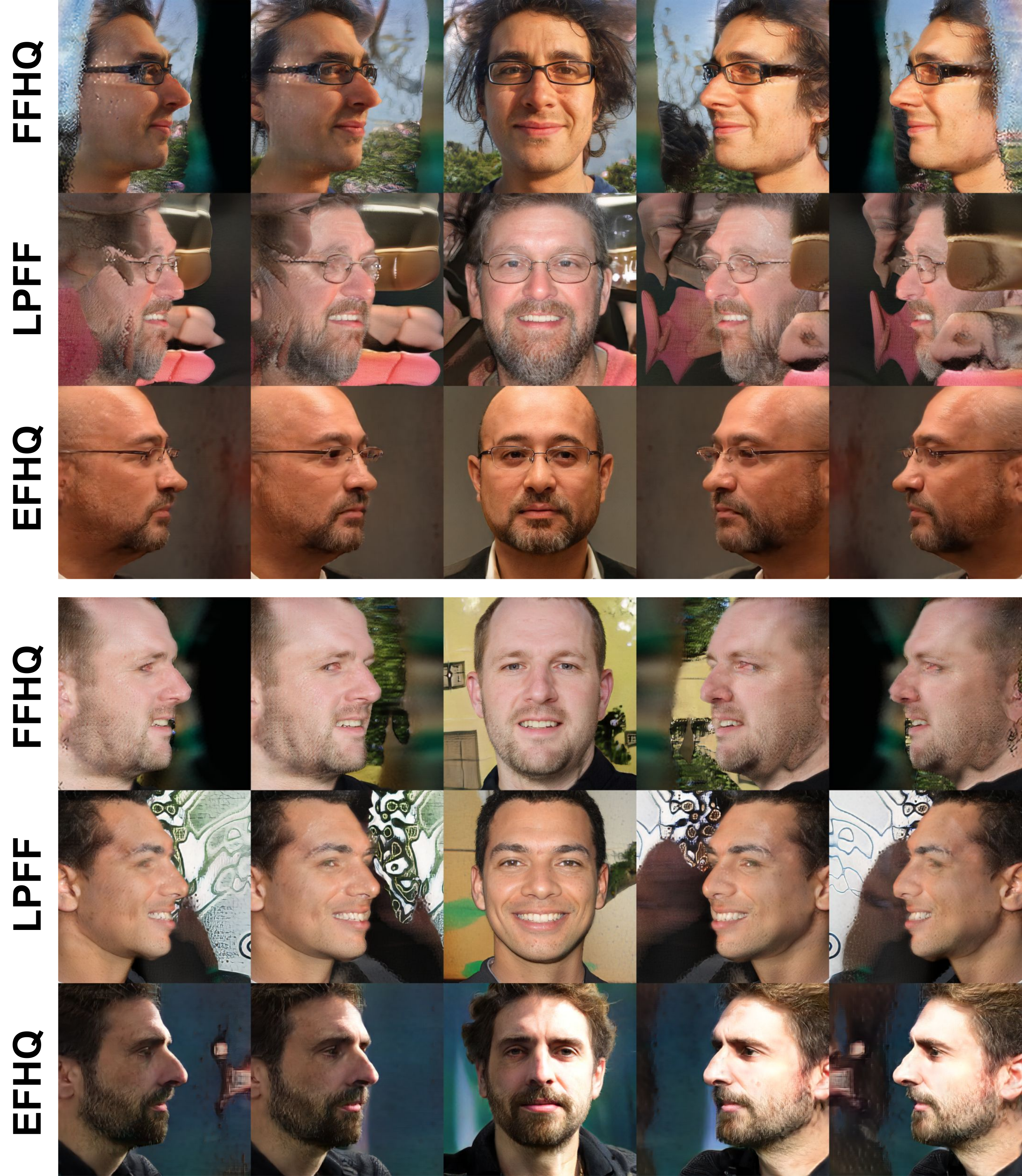}
    \vspace{-2mm}
\caption{\textbf{Comparison} between multiview generated samples, with truncation $\psi=0.8$, of EG3D model trained with various datasets (for both LPFF/EFHQ, the training dataset is combined with FFHQ).}
    \vspace{-2mm}
\label{fig:supp_eg3d4}
    \vspace{-4mm}
\end{figure*}

\begin{figure*}[ht!]
\centering
\includegraphics[keepaspectratio,width=0.95\linewidth]{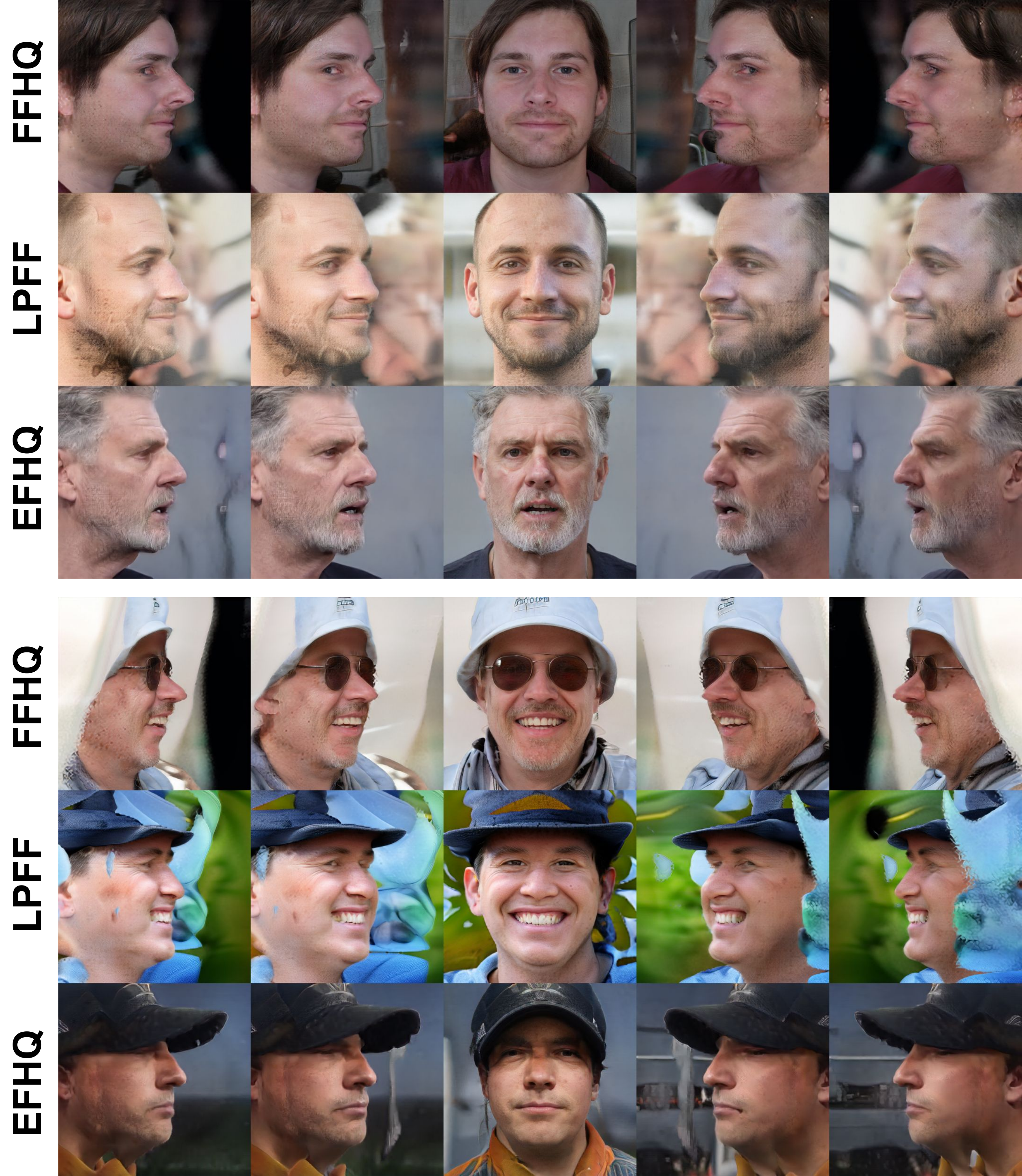}
    \vspace{-2mm}
\caption{\textbf{Comparison} between multiview generated samples, with truncation $\psi=0.8$, of EG3D model trained with various datasets (for both LPFF/EFHQ, the training dataset is combined with FFHQ).}
    \vspace{-2mm}
\label{fig:supp_eg3d5}
    \vspace{-4mm}
\end{figure*}

\begin{figure*}[ht!]
\centering
\includegraphics[keepaspectratio,width=\linewidth]{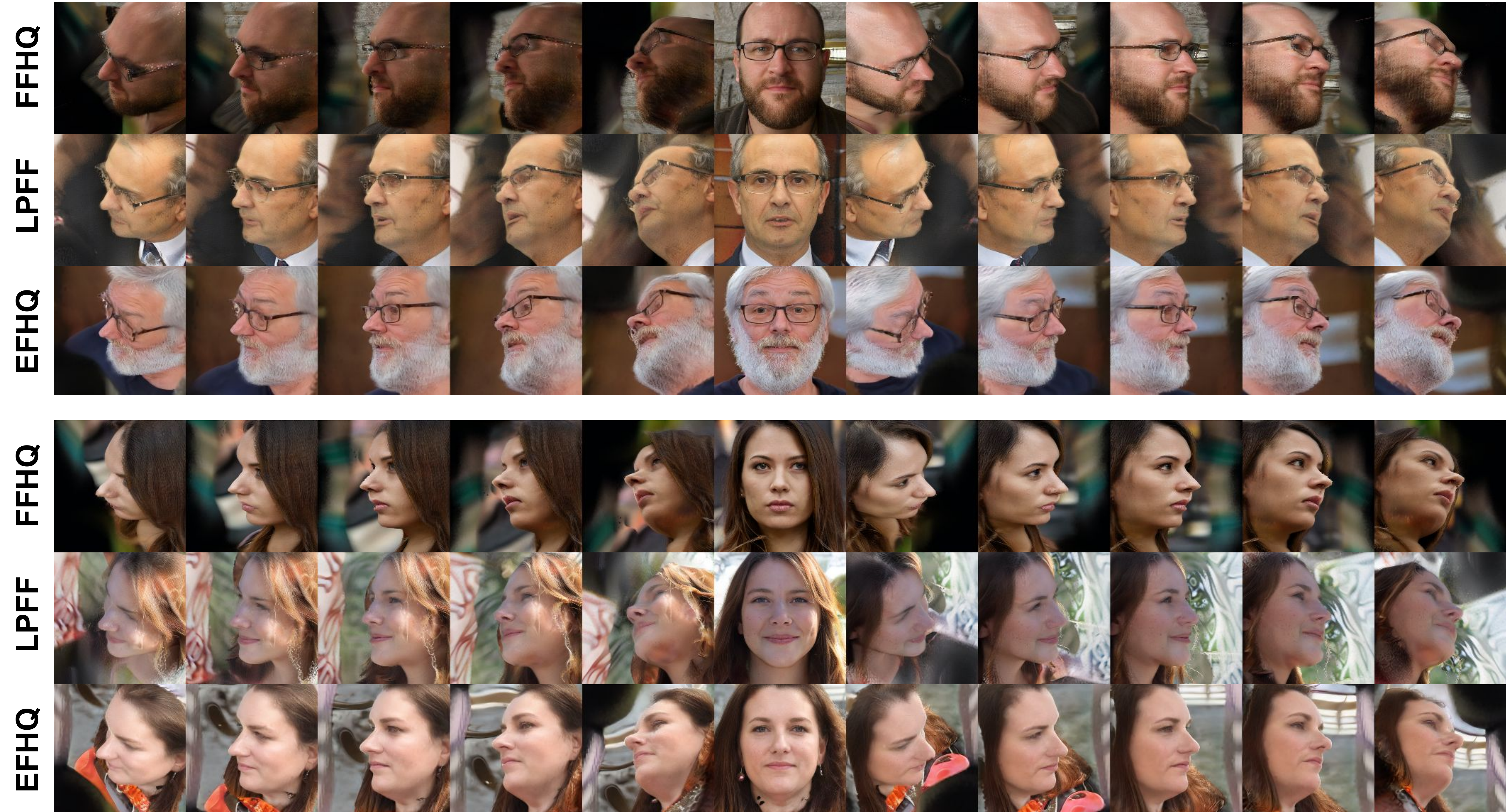}
    \vspace{-2mm}
\caption{\textbf{Comparison} between multiview generated samples, with truncation $\psi=0.8$, of EG3D model trained with various datasets (for both LPFF/EFHQ, the training dataset is combined with FFHQ).}
    \vspace{-2mm}
\label{fig:supp_eg3d_pitch1}
    \vspace{-4mm}
\end{figure*}

\begin{figure*}[ht!]
\centering
\includegraphics[keepaspectratio,width=\linewidth]{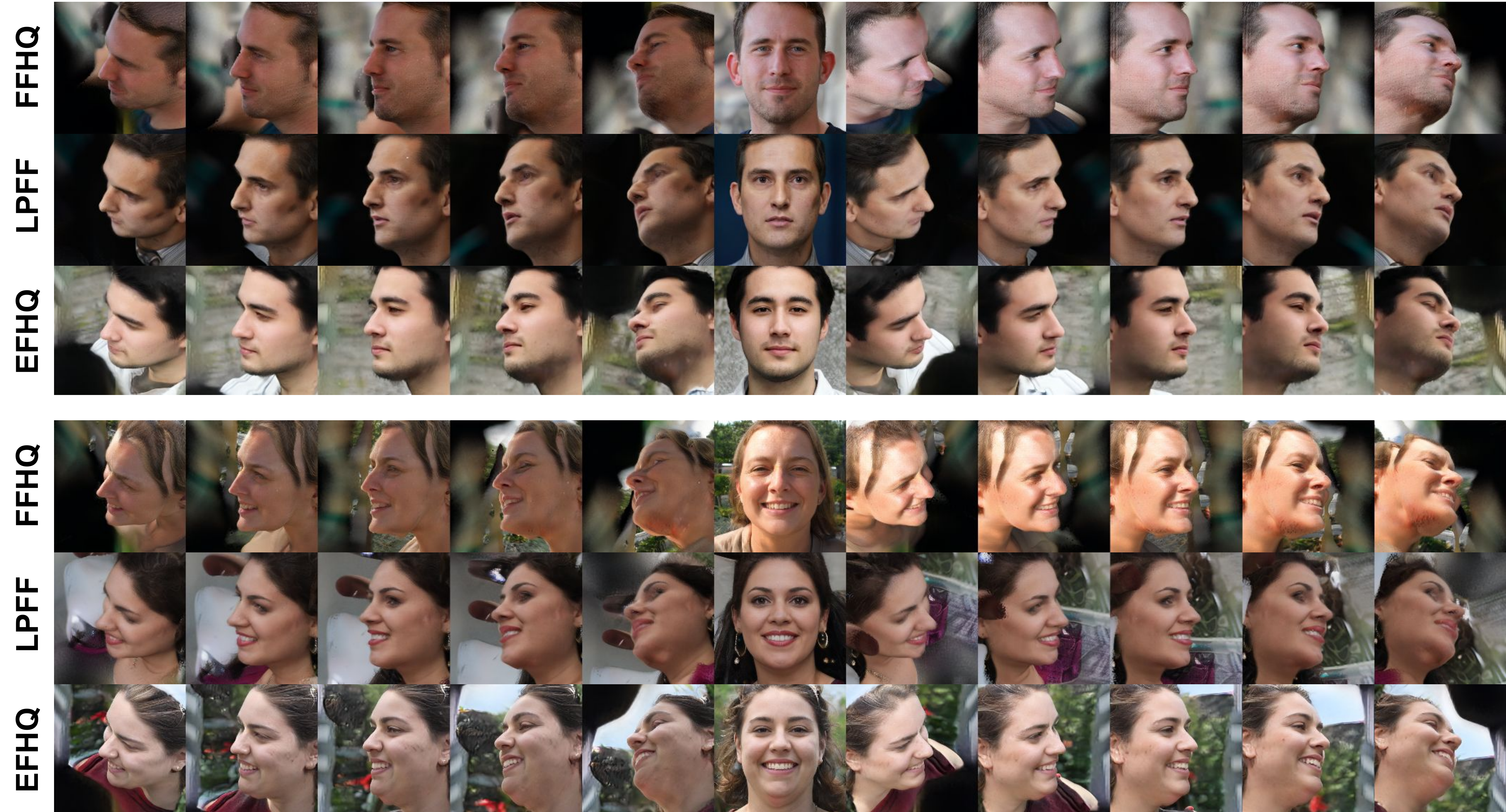}
    \vspace{-2mm}
\caption{\textbf{Comparison} between multiview generated samples, with truncation $\psi=0.8$, of EG3D model trained with various datasets (for both LPFF/EFHQ, the training dataset is combined with FFHQ).}
    \vspace{-2mm}
\label{fig:supp_eg3d_pitch2}
    \vspace{-4mm}
\end{figure*}

\begin{figure*}[ht!]
\centering
\includegraphics[keepaspectratio,width=0.95\linewidth]{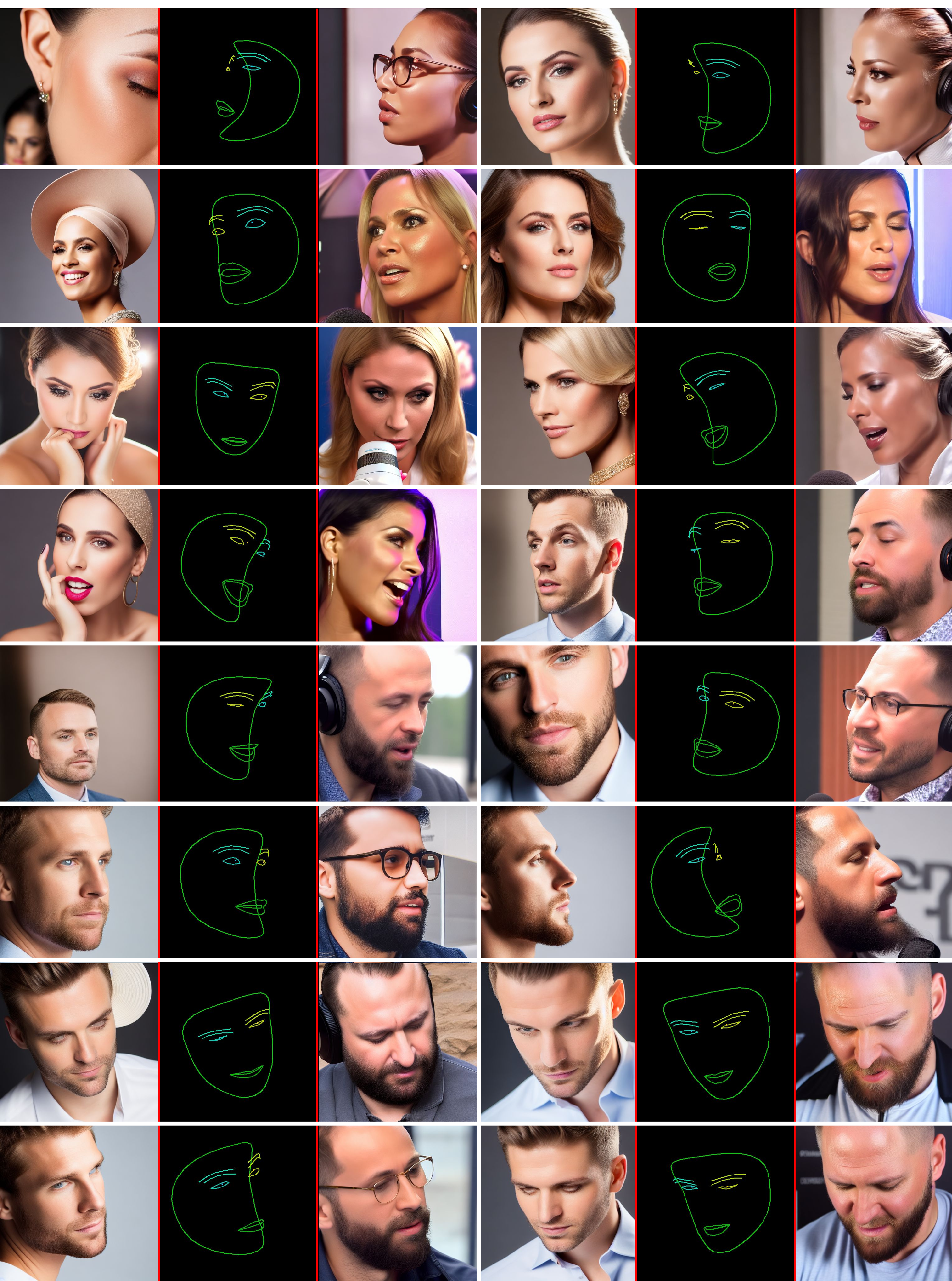}
    \vspace{-2mm}
\caption{\textbf{Comparison} profile-view generated samples of pretrained ControlNet (left) and our fine-tuned ControlNet (right) with the prompt: “A profile portrait image of a person.”}
    \vspace{-2mm}
\label{fig:supp_controlnet1}
    \vspace{-4mm}
\end{figure*}

\begin{figure*}[ht!]
\centering
\includegraphics[keepaspectratio,width=0.95\linewidth]{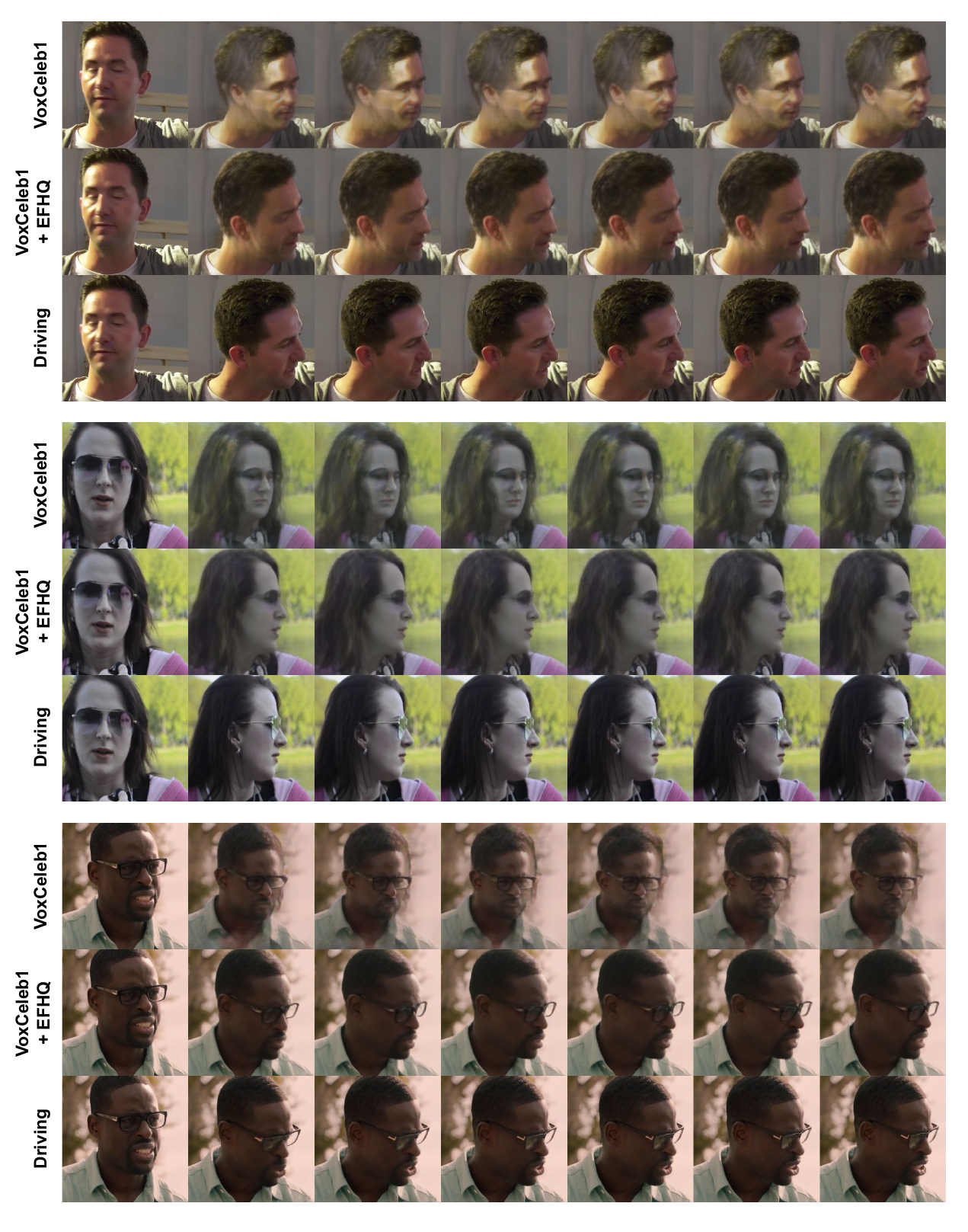}
    \vspace{-2mm}
\caption{\textbf{Comparison} between same-identity reenactment of TPS model trained with various datasets. The first frame of each row represents the source image, while the last row depicts the ground truth driving frames.}
    \vspace{-2mm}
\label{fig:supp_tps1}
    \vspace{-4mm}
\end{figure*}

\begin{figure*}[ht!]
\centering
\includegraphics[keepaspectratio,width=0.95\linewidth]{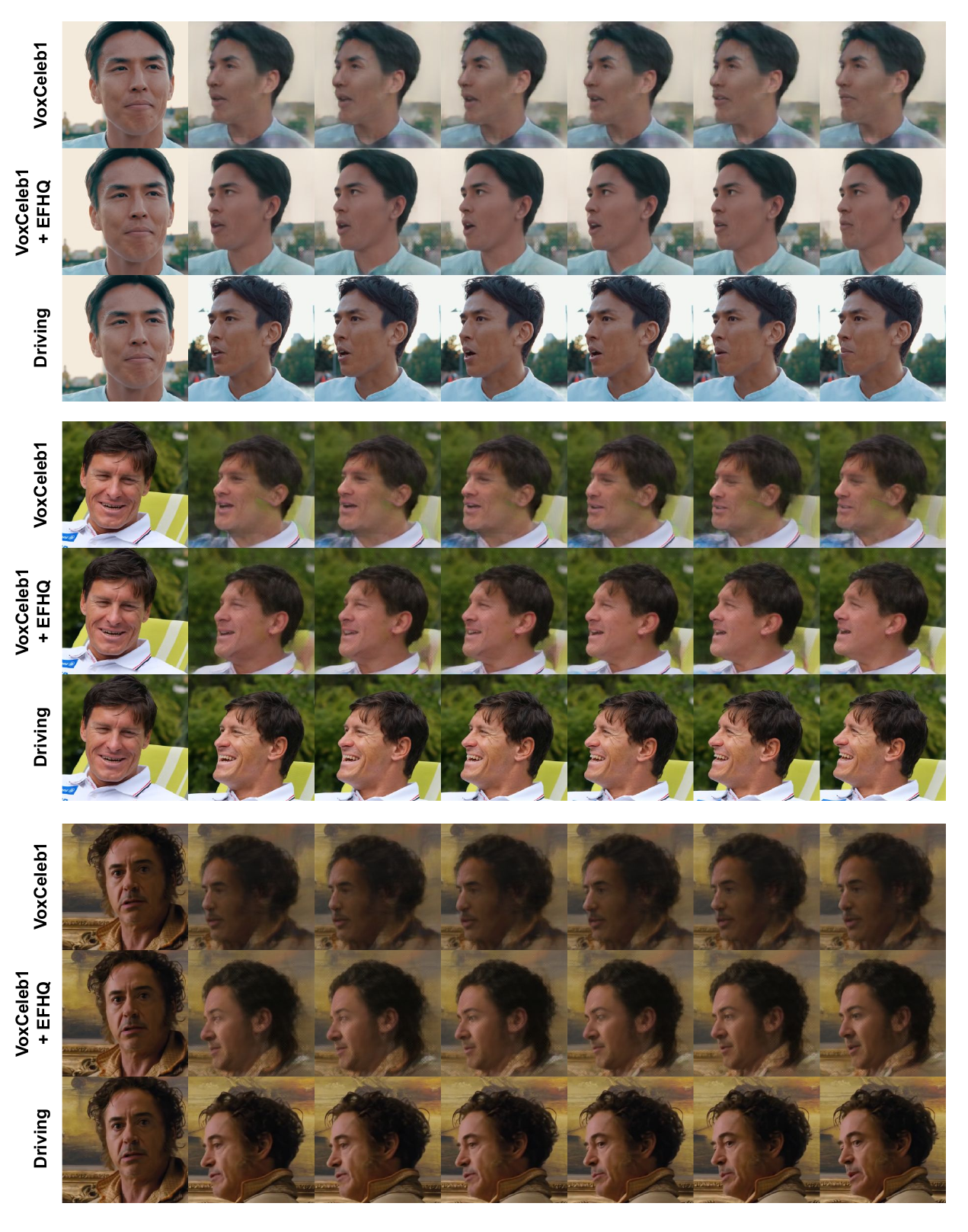}
    \vspace{-2mm}
\caption{\textbf{Comparison} between same-identity reenactment of TPS model trained with various datasets. The first frame of each row represents the source image, while the last row depicts the ground truth driving frames.}
    \vspace{-2mm}
\label{fig:supp_tps2}
    \vspace{-4mm}
\end{figure*}

\begin{figure*}[ht!]
\centering
\includegraphics[keepaspectratio,width=0.95\linewidth]{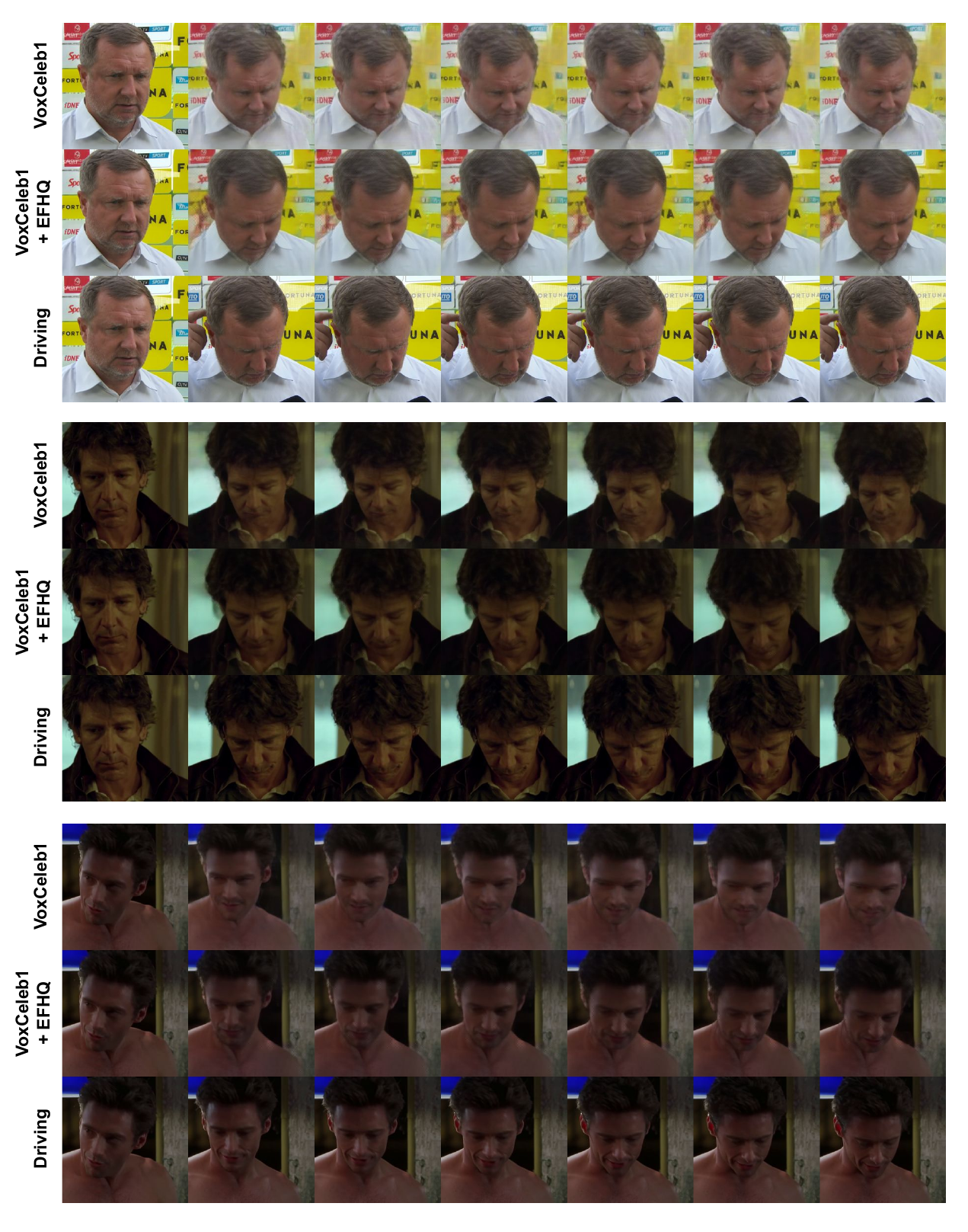}
    \vspace{-2mm}
\caption{\textbf{Comparison} between same-identity reenactment of TPS model trained with various datasets. The first frame of each row represents the source image, while the last row depicts the ground truth driving frames.}
    \vspace{-2mm}
\label{fig:supp_tps3}
    \vspace{-4mm}
\end{figure*}

\begin{figure*}[ht!]
\centering
\includegraphics[keepaspectratio,width=0.95\linewidth]{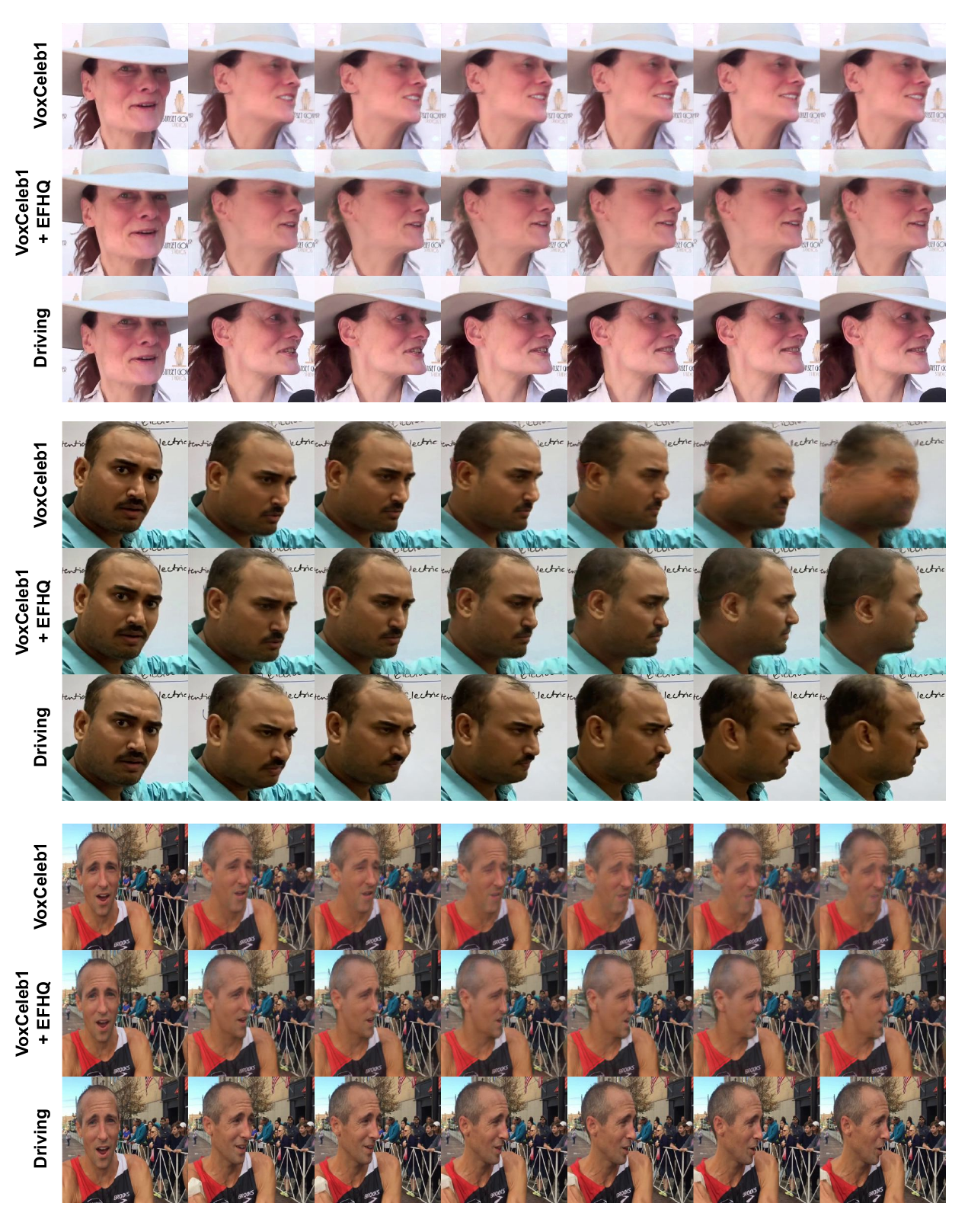}
    \vspace{-2mm}
\caption{\textbf{Comparison} between same-identity reenactment of LIA model trained with various datasets. The first frame of each row represents the source image, while the last row depicts the ground truth driving frames.}
    \vspace{-2mm}
\label{fig:supp_lia1}
    \vspace{-4mm}
\end{figure*}

\begin{figure*}[ht!]
\centering
\includegraphics[keepaspectratio,width=0.95\linewidth]{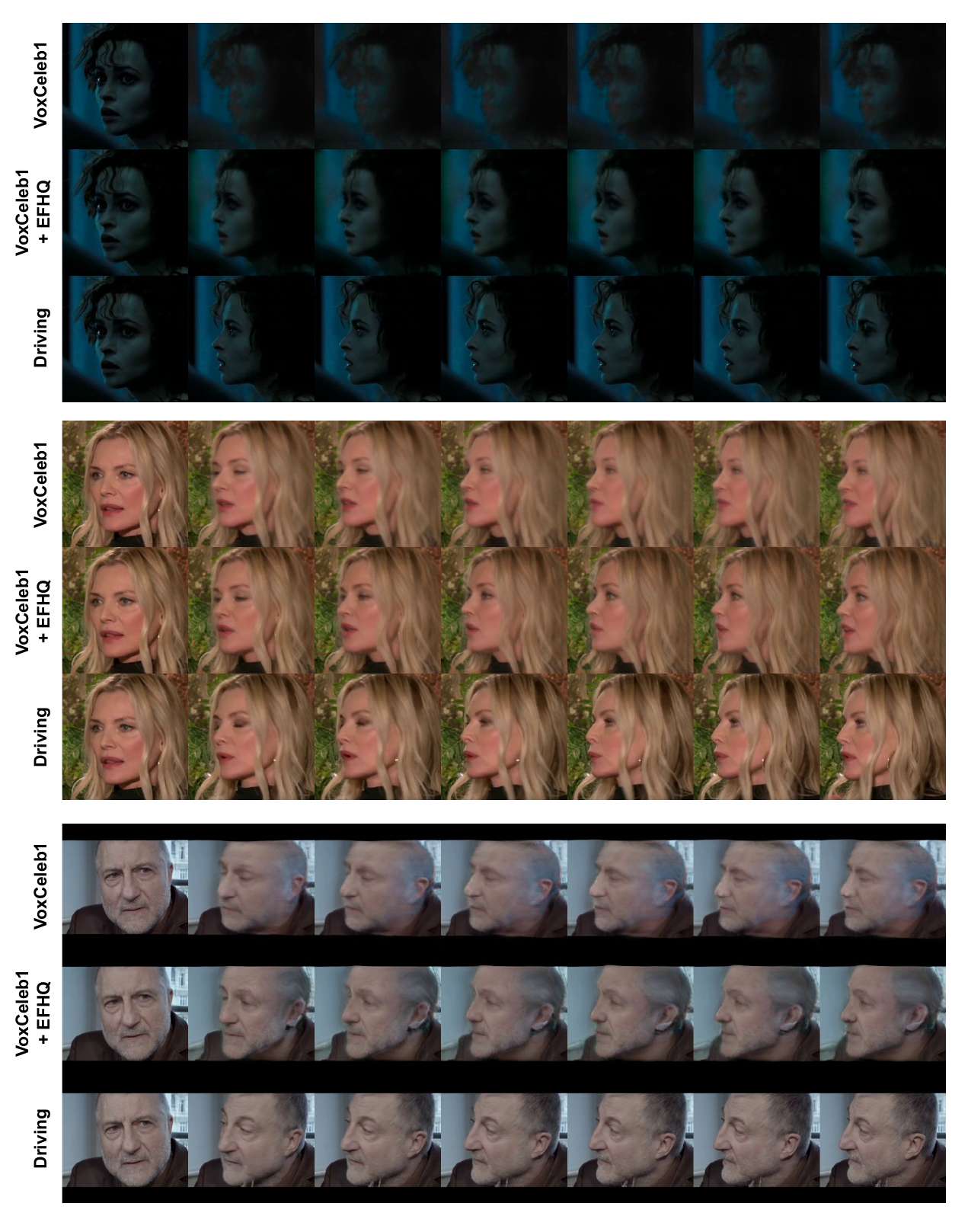}
    \vspace{-2mm}
\caption{\textbf{Comparison} between same-identity reenactment of LIA model trained with various datasets. The first frame of each row represents the source image, while the last row depicts the ground truth driving frames.}
    \vspace{-2mm}
\label{fig:supp_lia2}
    \vspace{-4mm}
\end{figure*}

\begin{figure*}[ht!]
\centering
\includegraphics[keepaspectratio,width=0.95\linewidth]{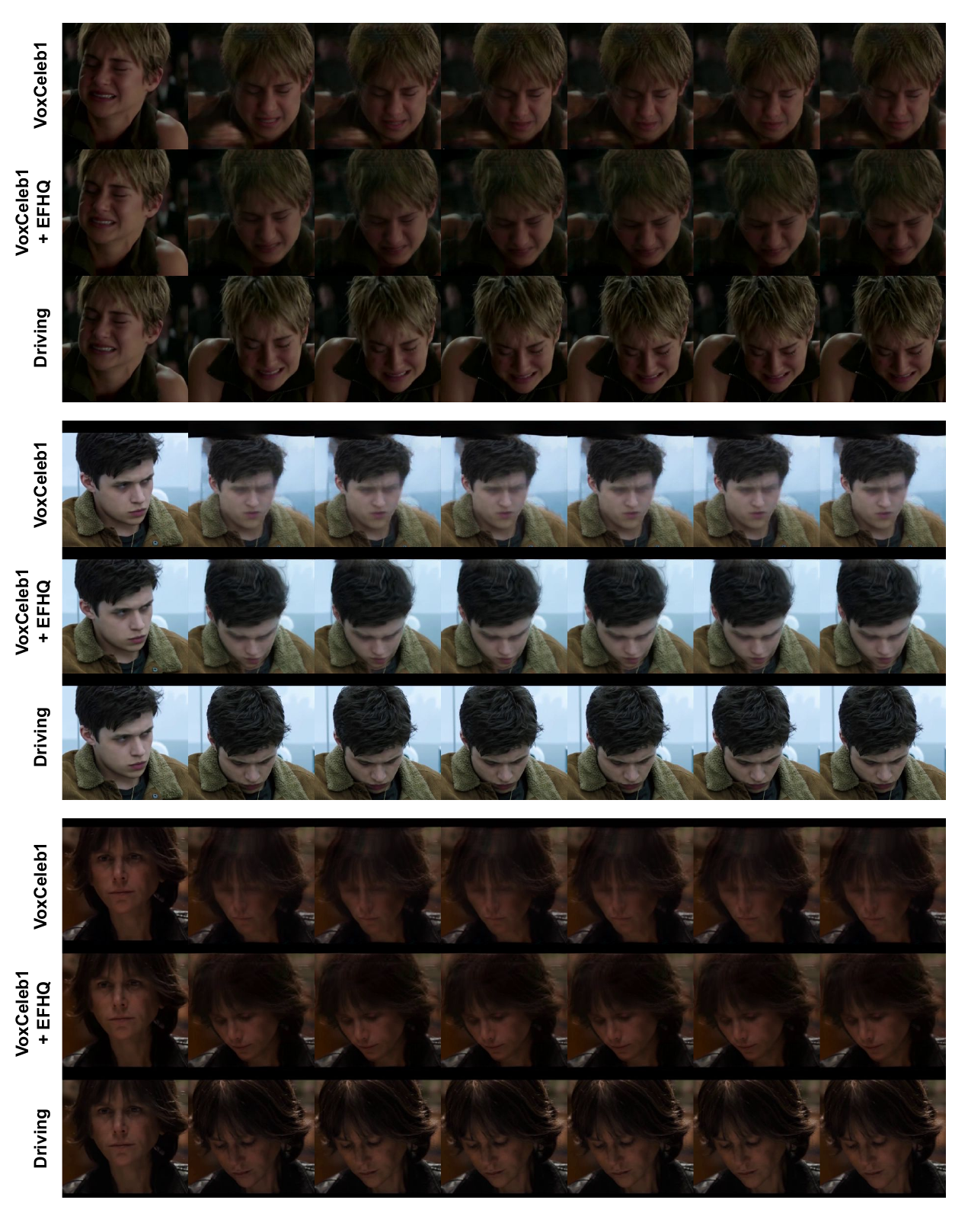}
    \vspace{-2mm}
\caption{\textbf{Comparison} between same-identity reenactment of LIA model trained with various datasets. The first frame of each row represents the source image, while the last row depicts the ground truth driving frames.}
    \vspace{-2mm}
\label{fig:supp_lia3}
    \vspace{-4mm}
\end{figure*}

\begin{figure*}[ht!]
\centering
\includegraphics[keepaspectratio,width=0.95\linewidth]{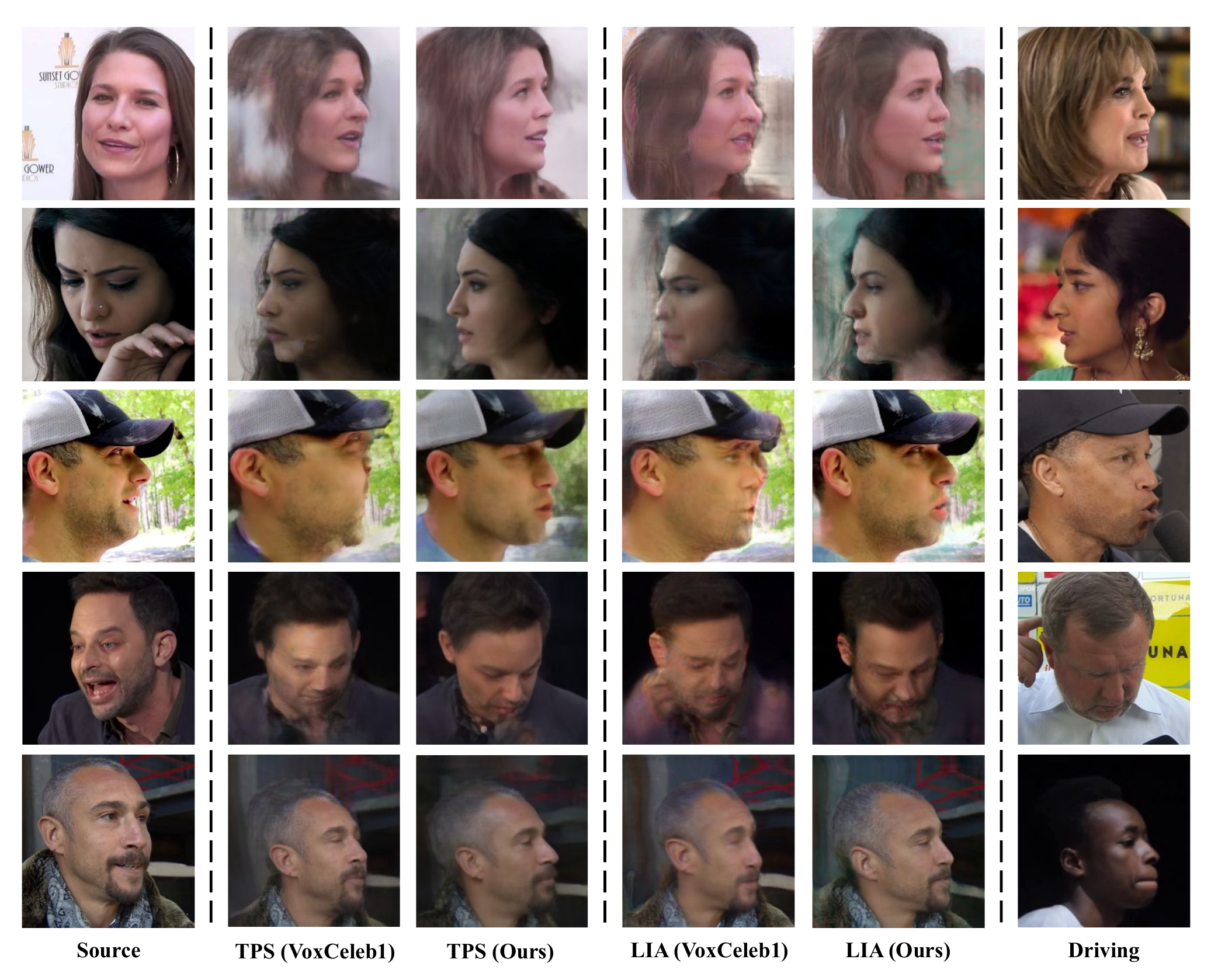}
    \vspace{-2mm}
\caption{\textbf{Comparison} between cross-identity reenactment of models trained with various datasets. The first frame of each row represents the source image, while the last frame depicts the ground truth driving frame.}
    \vspace{-2mm}
\label{fig:supp_cross_reenact}
    \vspace{-4mm}
\end{figure*}

We provide more qualitative examples of 2D/3D-aware GAN generation, diffusion-based text-to-image generation, and face reenactment to further demonstrate the superiority of our method. The short descriptions for the figures are shown below.
\begin{itemize}
    \item \cref{fig:supp_stylegan1} shows a comparative analysis conducted at the patch level between samples from StyleGAN2-ADA trained with FFHQ+LPFF and FFHQ+EFHQ datasets to further elucidate the quality and sharpness of the synthesized facial images. \cref{fig:supp_stylegan2} exhibits additional samples generated by our models.
    \item \cref{fig:supp_eg3d1,fig:supp_eg3d2,fig:supp_eg3d3,fig:supp_eg3d4,fig:supp_eg3d5} present a comparison of synthesized faces from various views produced by EG3D, trained with FFHQ, FFHQ+LPFF, and FFHQ+EFHQ datasets. Moreover, \cref{fig:supp_eg3d_pitch1,fig:supp_eg3d_pitch2} serve the same purpose as the aforementioned figures but with a different setup of viewing angles.
    \item \cref{fig:supp_controlnet1} present a comparison of synthesized faces from ControlNet trained with OpenPose\cite{openpose} 's dataset (released by ControlNet\cite{controlnet}) and trained with our dataset.
    \item \cref{fig:supp_tps1,fig:supp_tps2,fig:supp_tps3} present a comparison of same-identity face reenactment from TPS, trained with VoxCeleb1 and with VoxCeleb1+EFHQ.
    \item
    \cref{fig:supp_lia1,fig:supp_lia2,fig:supp_lia3} present a comparison of same-identity face reenactment from LIA, trained with VoxCeleb1 and with VoxCeleb1+EFHQ.
    \item
    \cref{fig:supp_cross_reenact} presents a comparison of cross-identity face reenactment from both LIA and TPS, trained with VoxCeleb1 and with VoxCeleb1+EFHQ.
\end{itemize}


\begin{figure*}[ht!]
\centering
\includegraphics[keepaspectratio,width=\linewidth]{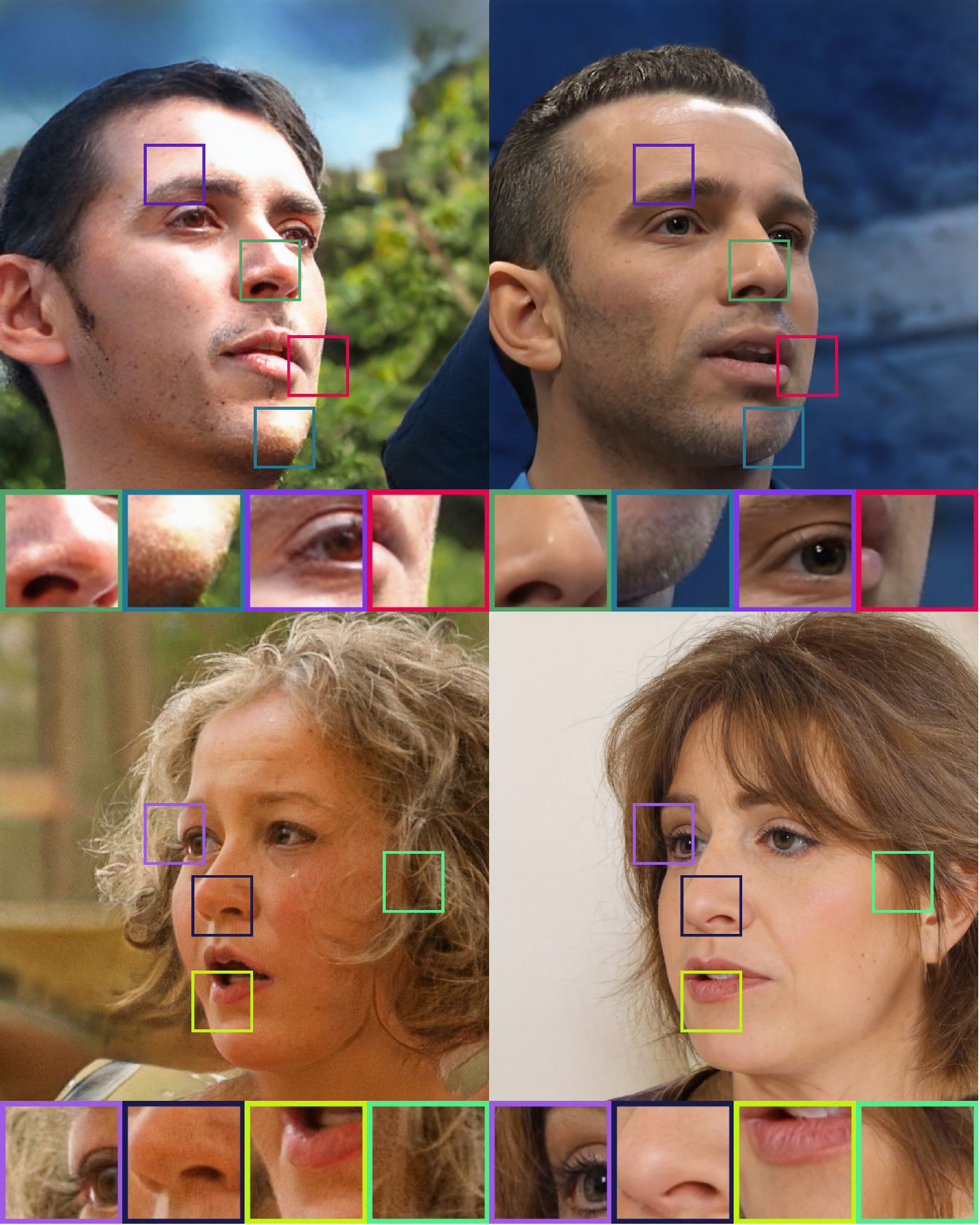}
\caption{Comparison between profile-view generated samples of StyleGAN2-ADA training with FFHQ+LPFF (left) and FFHQ+EFHQ (right), with truncation $\psi=0.7$.}
\label{fig:supp_stylegan1}
\end{figure*}



\end{document}